\documentclass{article}
\PassOptionsToPackage{numbers, compress}{natbib}
\usepackage[preprint]{neurips_2026}

\usepackage[utf8]{inputenc}
\usepackage[T1]{fontenc}
\usepackage[colorlinks=true,linkcolor=blue!70!black,citecolor=blue!70!black,urlcolor=blue!70!black]{hyperref}
\usepackage{url}
\usepackage{booktabs}
\usepackage{amsfonts}
\usepackage{nicefrac}
\usepackage{microtype}
\usepackage{xcolor}
\usepackage{graphicx}
\usepackage{subcaption}
\usepackage{amsmath}
\usepackage{amssymb}
\usepackage{amsthm}
\usepackage{algorithm}
\usepackage{algorithmic}
\usepackage{bm}
\usepackage{multirow}
\usepackage{placeins}

\hypersetup{
  pdftitle={Conditioning Protein Generation via Hopfield Pattern Multiplicity},
  pdfauthor={Jeffrey D. Varner},
  pdfsubject={Training-free conditioning of protein sequence generation},
  pdfkeywords={protein sequence generation, stochastic attention, modern Hopfield networks,
    profile hidden Markov models, Langevin dynamics, scarce-data protein design,
    conditional generation, protein engineering}
}

\newtheorem{theorem}{Theorem}

\newtheorem{proposition}[theorem]{Proposition}

\theoremstyle{definition}

\theoremstyle{remark}

\newcommand{\vxi}{\bm{\xi}}
\newcommand{\vX}{\mathbf{X}}
\newcommand{\vm}{\mathbf{m}}
\newcommand{\vr}{\mathbf{r}}
\newcommand{\veps}{\bm{\epsilon}}
\newcommand{\R}{\mathbb{R}}
\DeclareMathOperator{\softmax}{softmax}
\newcommand{\SIRef}[1]{Section~\ref{#1} of the Supporting Information}
\newcommand{\SIFigRef}[1]{Fig.~\ref{#1}}
\newcommand{\KunitzFobsFiveHundred}{0.587}
\newcommand{\KunitzFobsFiveHundredSD}{0.029}
\newcommand{\AttnMaxDevPts}{2.6}
\newcommand{\AttnMaxDevFamily}{$\omega$-Conotoxin}
\newcommand{\AttnMaxDevRho}{5}
\newcommand{\RelationIntercept}{0.73}
\newcommand{\RelationSlope}{-1.8}
\newcommand{\RelationRsq}{0.78}
\newcommand{\RelationLooSlopeMin}{-2.5}
\newcommand{\RelationLooSlopeMax}{-1.3}
\newcommand{\RelationLooRsqMin}{0.71}
\newcommand{\RelationLooRsqMax}{0.91}
\newcommand{\SARhoFiveHundredPPL}{4.28}
\newcommand{\SARhoFiveHundredPPLSD}{0.46}
\newcommand{\HMMRhoFiveHundredPPL}{6.54}
\newcommand{\HMMRhoFiveHundredPPLSD}{0.89}
\newcommand{\HMMRhoFiveHundredMarker}{0.996}
\newcommand{\HMMRhoFiveHundredMarkerSD}{0.002}
\newcommand{\HMMRhoFiveHundredKL}{5.6}
\newcommand{\HMMRhoFiveHundredKLSD}{0.4}
\newcommand{\HMMRhoFiveHundredDiversity}{0.590}

\newcommand{\KunitzSeparation}{0.20}

\newcommand{\SHThreeSeparation}{0.34}

\newcommand{\WWSeparation}{0.11}

\newcommand{\HomeoboxSeparation}{0.42}

\newcommand{\ForkheadSeparation}{0.17}

\newcommand{\ConotoxinSeparation}{0.78}

\title{Conditioning Protein Generation via Hopfield Pattern Multiplicity}

\author{%
  Jeffrey D.~Varner \\
  Robert Frederick Smith School of Chemical and Biomolecular Engineering\\
  Cornell University, Ithaca, NY 14850 \\
  \texttt{jdv27@cornell.edu} \\
}

\begin{document}

\maketitle

\begin{abstract}
Small protein-family alignments often contain a subset of interest but not enough labeled data
to train a conditional generator. We condition a training-free stochastic-attention sampler by
adding one multiplicity ratio to its logits. Increasing this ratio shifts generation from the
full family toward the designated subset. For unit-norm memories, the resulting Boltzmann
distribution is exactly a Gaussian mixture whose component weights are set by the
multiplicities. This result separates exact conditioning in latent space from losses caused by
sampling, PCA reconstruction, and sequence decoding. Across five Pfam families, attention
followed the analytic target, but recovery of single-residue markers depended on how well PCA
separated the designated and background sequences. A matched weighted profile HMM reproduced
these markers more directly, while stochastic attention gave lower ESM2 pseudo-perplexity in
the Kunitz comparison. Using a curated set of 23 $\omega$-conotoxin sequences as the target subset
produced diverse sequences that preserved the cysteine scaffold and Tyr13 and shifted other
residues toward the designated set. These sequences are candidates for experimental testing;
they do not establish binding.

\end{abstract}

\section{Introduction}
Generating novel protein sequences that fold correctly and perform a desired function is a
central goal of computational protein engineering. Several classes of methods now address this
problem. Deep generative models, including variational autoencoders~\cite{hawkinsHookerVAE2021},
autoregressive language models~\cite{madaniProGen2023}, and diffusion
models~\cite{alamdariProteinGeneration2023}, learn sequence distributions from large databases
and can produce diverse candidates, but require substantial training data and GPU resources.
Structure-conditioned inverse folding methods~\cite{dauparasProteinMPNN2022} design sequences for
a given backbone but require a target structure as input. Protein language
models~\cite{esmfold2022} trained on evolutionary databases can score or sample sequences, but
conditioning them on a specific functional property typically requires fine-tuning on labeled
data. Profile hidden Markov models (HMMs)~\cite{hmmer3} provide a standard training-free model
of family alignments. Sequence weights can shift their position-specific emission marginals
toward a designated subset, making a weighted profile HMM a direct comparator for the present
method. Profile HMMs do not model correlations between match-state emissions, whereas
stochastic attention samples a continuous representation of complete stored sequences. We
therefore compared the two methods under the same alignments, designation labels, and multiplicity ratios. For the
scarce-data regime considered here (families with tens to hundreds of members, typical of Pfam
seed alignments), models trained on the target family alone have limited training signal.
Pretrained models are not restricted in this way, since they draw on sequence data beyond the
target family.

Stochastic attention (SA) was developed for the scarce-data regime where traditional methods struggle due to limited training data~\cite{varnerSAProtein2026}. Stochastic attention treats the energy function of a modern Hopfield network, constructed directly from a small family alignment, as a Boltzmann density and samples from it using Langevin dynamics. The method requires no training, runs on a laptop, and produces sequences that preserve family-level amino acid composition, per-position conservation, and predicted three-dimensional fold. A key
limitation, however, is that SA treats all sequences in the alignment equally. Every family
member contributes the same weight to the energy landscape, so the sampler explores the full
family distribution without preference for any functional subset. In practice, experimentalists
often want something more specific: not just a plausible Kunitz domain, but a Kunitz domain
that \emph{inhibits trypsin}; not just a plausible SH3 domain, but one that binds a
particular proline-rich motif. The members of a family alignment share a fold but differ in
function, and the residues that carry the difference occupy only a few positions. A sampler
that reproduces the family distribution faithfully therefore reproduces its variation at those
positions, producing the functionally relevant residue only as often as it appears in the
alignment. This raises the question of how to condition the generative process on a functional
subset without retraining.

In this study, we answered that question with multiplicity weighting. Stochastic attention
encodes each aligned sequence as a column in a memory matrix. At each step the sampler
computes attention weights over these stored sequences and moves toward their weighted average.
A bias added to the attention scores makes it spend more time near selected sequences. We
assigned a multiplicity weight $r_k > 0$ to each stored sequence and added $\log r_k$ to its
attention score before the softmax normalization. All designated sequences share one weight and
all background sequences share another. A single number, the multiplicity ratio $\rho$,
therefore controls the bias: $\rho = 1$ recovers standard SA, while increasing $\rho$ shifts
generation toward the designated subset.
The method is agnostic to what the designated subset represents (binding, thermostability, organism of origin); it simply biases the sampler toward the sequences the user points to. 
The same multiplicity-weighted mechanism has also been applied to synthetic patient generation
from small longitudinal clinical cohorts~\cite{varnerSAPatient2026}, so the conditioning is not
specific to sequence data. We compared multiplicity weighting against the obvious alternative,
hard curation, which builds the memory matrix from the designated sequences alone and leaves
the rest of the family out. Those discarded sequences are what show which
positions the fold holds fixed and which are free to vary, so dropping them leaves the sampler
with far fewer examples to work from. We tested both on five
Pfam protein families (Kunitz, SH3, WW, Homeobox, and Forkhead domains) spanning a range of
family sizes ($K = 55$--$420$) and designated-subset geometries. In Kunitz, hard curation
retained the selected marker from as few as three designated inputs. Those inputs were chosen
because they carry the marker, so recovery demonstrated control over the sampler, not the
function associated with that marker. The multiplicity weights,
by contrast, enter the sampler as a simple logit bias, preserving the analytic score function and
$\mathcal{O}(dK)$ per-step cost of the original method. At ideal equilibrium, $\rho$ sets the
latent mixture share exactly. In the finite replicated runs, mean attention on designated
patterns remained within \AttnMaxDevPts{} percentage points of that share at worst
(\AttnMaxDevFamily{}, $\rho=\AttnMaxDevRho$). The discrete outputs were less faithful.
Finite-temperature sampling, reconstruction from unit-normalized principal component analysis
(PCA) coordinates, and argmax decoding could leave the marker frequency below the designated
mixture share. We called this difference the calibration gap when designation and marker were
the same label. We then compared it with the separation index $S$, a simple measure of how well
PCA separates designated and background sequences. Multiplicity weighting therefore offers a
practical route to scarce-data conditioning, but its decoded output must be calibrated.

\section{Results}

We evaluated two forms of training-free conditioning across six protein families
(Table~\ref{tab:cross-family}). Kunitz domains served as the primary test case, and four
additional Pfam seed families---SH3, WW, Homeobox, and Forkhead---tested whether the findings
generalized across family size and PCA geometry. The $\omega$-conotoxin O-superfamily provided
a distinct peptide example associated with the Cav2.2 calcium
channel~\cite{omegaConotoxinSAR2006}. In every family, we
compared hard curation, which rebuilds the memory from a designated subset, with multiplicity
weighting, which retains the full-family memory but increases the designated patterns' weight.
No activity measurements were available for these families, so every designated subset was
defined from sequence alone: the sequences carrying a chosen residue class at one alignment
column for the Pfam families, and a curated list of accessions for the $\omega$-conotoxins.
Each designation was therefore a sequence feature and not a measured function.

\begin{figure}[p]
  \centering
  \includegraphics[width=\textwidth]{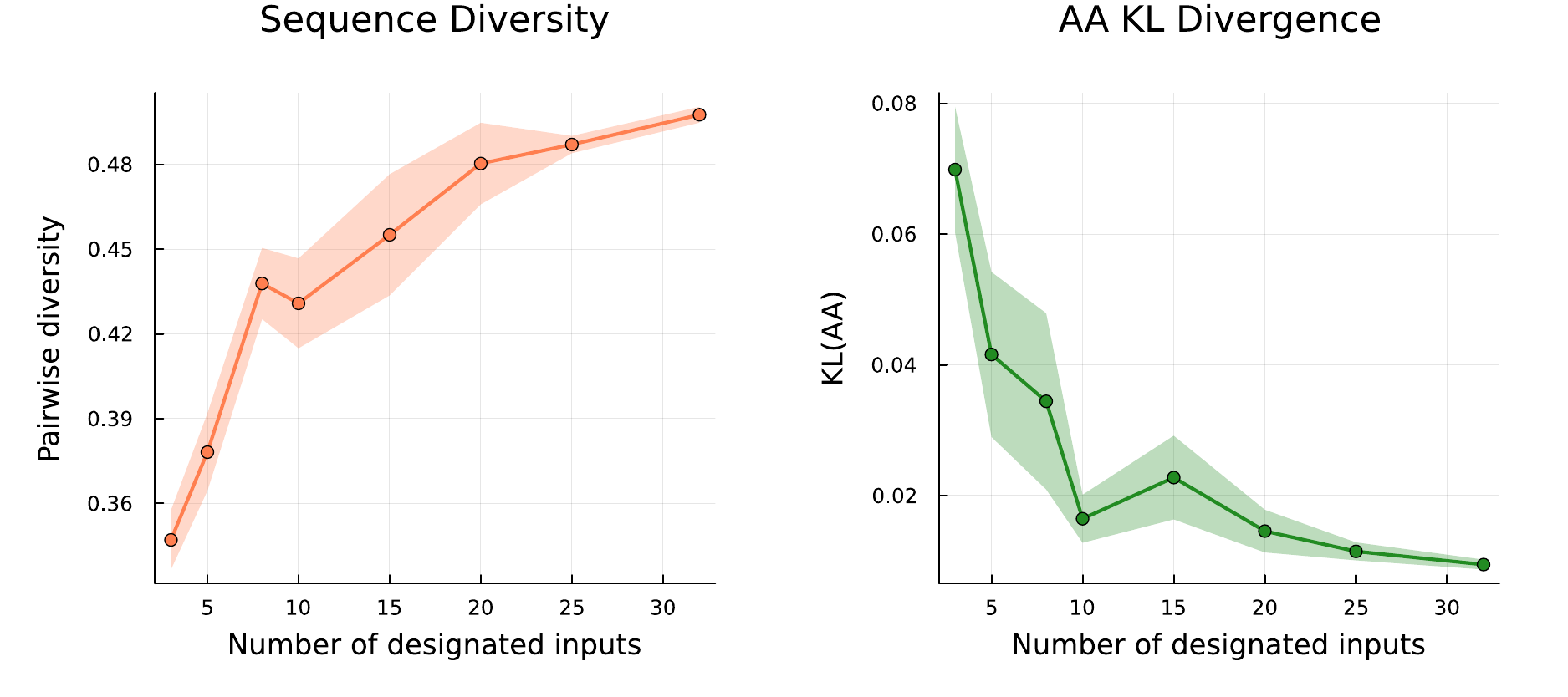}
  \caption{\textbf{Sequence properties vary with the number of designated inputs.}
  Subsampling $K_{\mathrm{des}} \in \{3,5,8,10,15,20,25,32\}$ sequences from
  the Kunitz K/R-positive set (3 replicates each; shaded bands show $\pm 1$
  s.d.). P1 K/R fraction was 1.0 at every sample size (not shown), so every input
  size retained the selected marker.
  \textit{Left:} Pairwise sequence diversity increases with the number of
  designated inputs. \textit{Right:} KL divergence of amino acid composition
  decreases over the same range.}
  \label{fig:scaling}
\end{figure}

\begin{figure}[p]
  \centering
  \includegraphics[width=\textwidth]{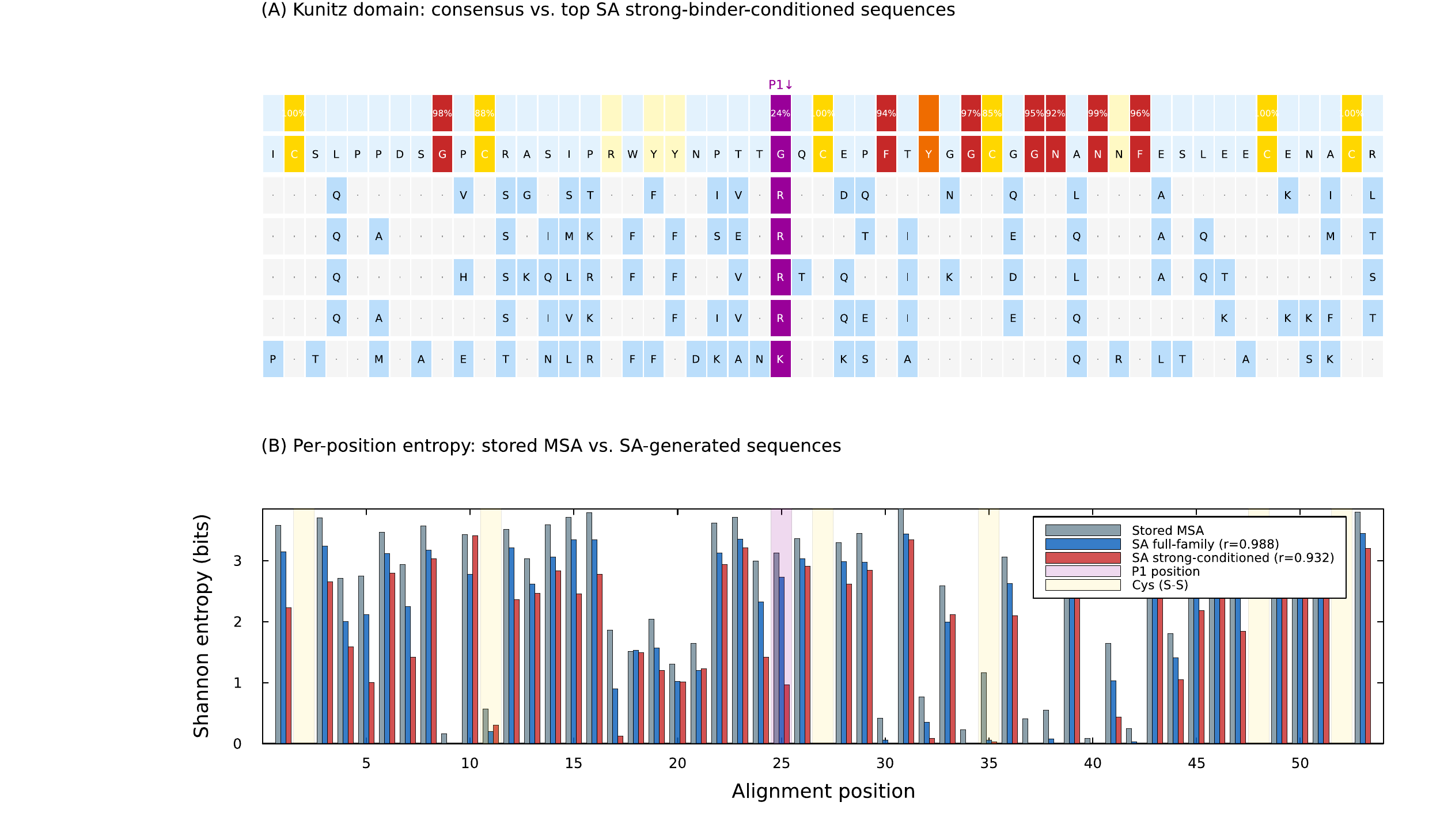}
  \caption{\textbf{Sequence-level analysis of SA-generated Kunitz domain sequences
  conditioned on the K/R-positive subset.}
  \textbf{(A)}~Alignment of the family consensus with the five highest-confidence
  SA K/R-positive-subset-conditioned sequences (ranked by ESMFold pLDDT). Positions are
  colored by conservation in the stored MSA: highly conserved ($>90\%$, red),
  conserved ($70$--$90\%$, orange), moderate ($50$--$70\%$, yellow), and variable
  ($<50\%$, blue). Cysteines forming the three disulfide bonds are highlighted in
  gold; the P1 position is highlighted in purple. Dots indicate matches to the
  consensus; letters indicate substitutions. All five sequences preserve every
  highly conserved position ($11/11$), all six cysteines ($6/6$), and carry K or R
  at P1, while introducing $19$--$26$ substitutions concentrated at variable sites.
  \textbf{(B)}~Per-position Shannon entropy for the stored MSA (gray), SA full-family
  generation (blue), and SA K/R-positive-subset-conditioned generation (red).
  Full-family generation tracks the stored entropy profile closely, showing similar
  position-specific conservation in the generated and stored sequences. The
  K/R-positive-subset profile tracks it slightly less closely, with the largest
  deviation at the P1 position (column~25): conditioning collapses P1 diversity to
  K/R, reducing entropy at this site while preserving the entropy profile
  elsewhere.}
  \label{fig:sequence-analysis}
\end{figure}

We first asked whether hard curation could transfer a selected marker while preserving the
broader family pattern (Fig.~\ref{fig:scaling}). Kunitz domains (PF00014; $K=99$, $L=53$ aligned positions) provided an
interpretable test: their P1 residue influences which protease they inhibit, and Lys or Arg at
P1 is associated with trypsin specificity~\cite{laskowskiProteinInhibitors1980}. We designated the
32 K/R-at-P1 sequences as marker-positive and the remaining 67 as marker-negative.
Separate memories built from these two subsets produced the sharp result expected when the
input was selected on the marker itself. All 930 sequences generated from the marker-positive
memory carried K/R at P1, and none generated from the marker-negative memory did; full-family
generation produced 38\% K/R compared with 32\% in the input family. The conditioning remained
localized rather than erasing the family pattern. The top marker-positive sequences retained
all 11 highly conserved positions and all six cysteines
while carrying 19--26 substitutions concentrated at variable sites. Per-position Shannon
entropy correlated with that of the stored multiple sequence alignment (MSA) for both full-family
generation ($r=0.988$) and marker-positive conditioning ($r=0.932$), with the largest change
at P1 (Fig.~\ref{fig:sequence-analysis}). The result also held when the designated input was
subsampled from 32 sequences down to three: P1 retention remained 100\% at every tested size,
whereas pairwise diversity rose and Kullback--Leibler (KL) divergence fell as more inputs were
added (Fig.~\ref{fig:scaling}). Thus, even three inputs were sufficient to fix the selected
marker, while additional inputs coincided with a broader PCA subspace and better distributional fit.

While hard curation provides a clear endpoint, it discards the background sequences and changes
the PCA basis. We therefore asked whether multiplicity weighting could tune Kunitz generation
continuously on the full-family basis (Fig.~\ref{fig:phase-transition}). The canonical
replicated analysis swept the multiplicity ratio $\rho$ between designated and background
weights from 1 (no bias) to 500 (strong bias); at each value, we recomputed the entropy-crossover temperature
$\beta^{*}$ and generated five independent libraries of 620 sequences (Table~\ref{tab:per-family-rho}).
We then compared the attention the sampler placed on designated patterns with the fraction of
decoded sequences carrying the P1 marker. Raising $\rho$ moved the attention, but the decoded
fraction followed only partway. At $\rho=500$, the weights called for 99.6\% of the sampler's
equilibrium mixture to sit on the designated Kunitz patterns. Every designated pattern carries
K/R at P1 and no background pattern does, so a library that carried the weighting through
unchanged would show K/R at P1 in 99.6\% of its sequences. The decoded fraction reached only
\KunitzFobsFiveHundred{} $\pm$ \KunitzFobsFiveHundredSD{}. This discrepancy between the target
marker fraction and the decoded one is the calibration gap $\Delta$
(Eq.~\ref{eq:decoder-pushforward}).
Increasing $\rho$ also produced modest global changes:
sequence diversity decreased from 0.56 to 0.51, KL divergence increased from 0.007 to 0.017,
and $\beta^{*}$ shifted from 4.4 at $\rho=1$ to 9.3 at $\rho=1{,}000$. Over that range the
effective pattern count, a measure of how concentrated the multiplicity
weights have become, fell from 99 to 32.1, approaching the concentration expected when weight
is confined to the 32 designated patterns. The decoded marker fraction nevertheless stayed far
below its target.

To locate the shortfall, we compared the analytic designated mixture share, empirical attention,
and decoded marker frequency. Across 17 values of $\rho$ from 1 to 1{,}000, mean designated
attention remained within 1.2 percentage points of the analytic share. The marker shortfall
therefore appeared downstream of latent weighting, where finite-temperature variation,
unit-normalized PCA reconstruction, and argmax decoding act together. Designated and background Kunitz sequences overlap in the
80-dimensional PCA basis (separation index $S=\KunitzSeparation$), so a decoded sequence need
not identify the component that generated its latent state. An exact-equilibrium comparison also
indicated that finite-step sampling was not the main source of the shortfall. At $\rho = 1$, 10,
and 500, we compared 1{,}640 independent draws from the
closed-form Gaussian mixture model with 1{,}640 retained unadjusted Langevin (ULA) states under
the same $\beta^{*}$ and decoder. At $\rho=500$, exact and ULA sampling produced nearly
identical decoded P1~K/R fractions of 0.604 and 0.597, and their decoded amino acid compositions differed by a KL
divergence of only $1.19\times10^{-4}$. Removing finite-step sampling error therefore moved the
decoded fraction by 0.007, against a shortfall of 0.39 from the 0.996 target. At all three values
of $\rho$, the fraction of states whose most probable mixture component was a designated
pattern differed by at most 0.026 between the two samplers, and the decoded marker fractions by
at most 0.016. The calibration gap therefore persisted at exact equilibrium and arose mainly
after latent sampling, in the combined reconstruction and decoding path.

\begin{figure}[p]
  \centering
  \includegraphics[width=\textwidth]{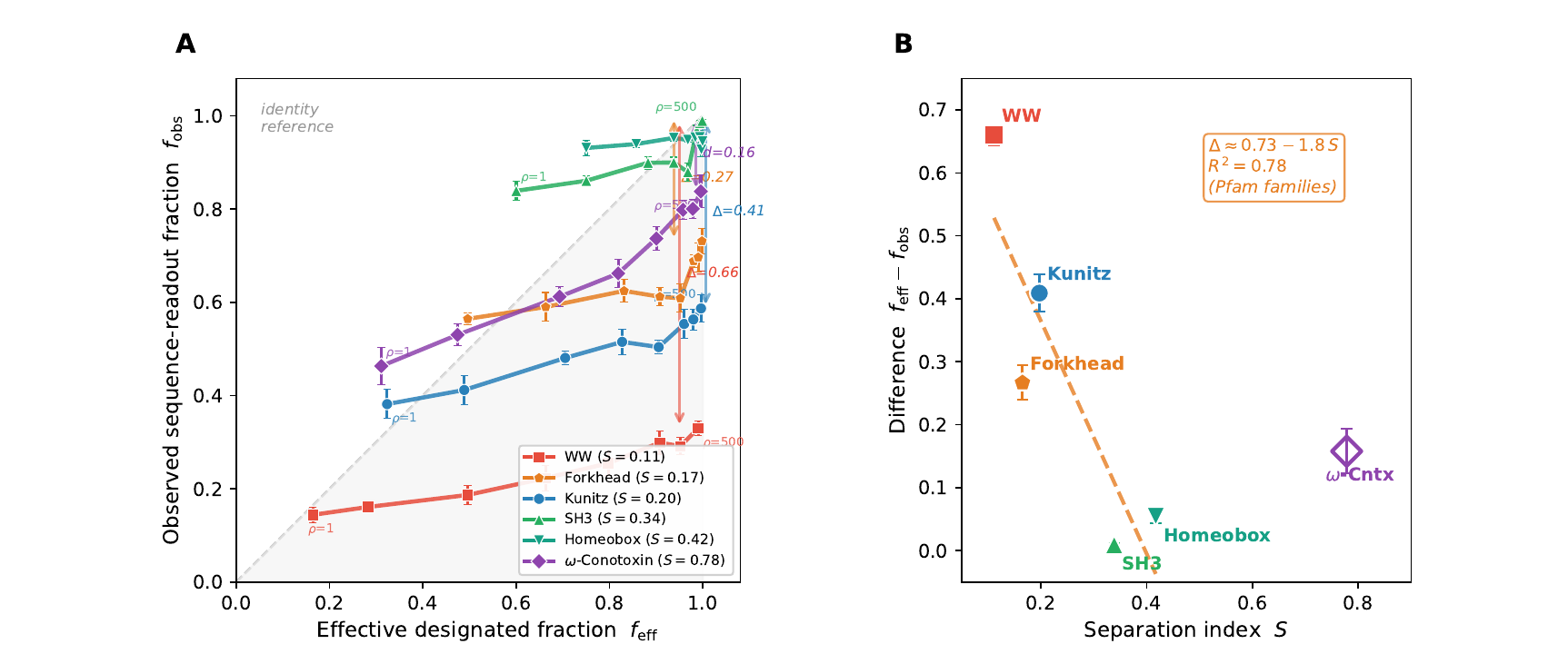}
  \caption{\textbf{Decoded sequence readouts vary with PCA-space separation.}
  \textbf{(A)}~Observed readout fraction $f_{\mathrm{obs}}$ versus effective designated
  fraction $f_{\mathrm{eff}}$ as the multiplicity ratio $\rho$ increases from 1 to 500
  for six protein families. The dashed diagonal marks perfect calibration
  ($f_{\mathrm{obs}} = f_{\mathrm{eff}}$). Homeobox
  ($S = \HomeoboxSeparation$) and SH3 ($S = \SHThreeSeparation$)
  domains track the diagonal closely; Kunitz ($S = \KunitzSeparation$) and Forkhead
  ($S = \ForkheadSeparation$) curve below; WW domains ($S = \WWSeparation$) show the
  largest departure. The $\omega$-conotoxin curve compares accession-based designation
  with a Tyr13 readout rather than like quantities. For the five Pfam families,
  double-headed arrows indicate the calibration gap at $\rho = 500$.
  \textbf{(B)}~Separation index $S$ versus calibration gap $\Delta$ at $\rho = 500$.
  Error bars show replicate s.d. An exploratory linear fit to the five Pfam families
  (filled markers; $\Delta \approx \RelationIntercept{} \RelationSlope\,S$,
  $R^{2}=\RelationRsq$) summarizes the association without defining a prospective
  threshold. The $\omega$-conotoxin point (open diamond) is displayed as an external
  comparison and is not included in the fit; its vertical difference is descriptive,
  not a calibration error.}
  \label{fig:separation-vs-gap}
\end{figure}

The Kunitz result suggested that conditioning should transfer most effectively when the PCA
representation separates the designated and background subsets. We tested this hypothesis on
four additional Pfam families, tracking Trp at the peptide-groove column for SH3 (PF00018;
$K=55$), a specificity-loop residue for WW (PF00397; $K=420$), Gln at position~50 of the
recognition helix for Homeobox (PF00046; $K=136$), and His or Asn at the H3 recognition-helix
position for Forkhead (PF00250; $K=246$). Together, the five Pfam splits sampled distinct
PCA geometries at comparable conditioning strengths (Table~\ref{tab:per-family-rho}). The
separation index, larger when PCA resolves the two subsets more cleanly, ranged from
\WWSeparation{} for WW to \HomeoboxSeparation{} for Homeobox.
The attention tracked its target across the panel: over all six replicated family
sweeps, the largest absolute deviation between the attention on designated patterns and the
analytic designated share was \AttnMaxDevPts{} percentage points (\AttnMaxDevFamily{},
$\rho=\AttnMaxDevRho$). What varied was the decoded marker response.
Calibration gaps decreased with separation
overall, although not monotonically, and an exploratory five-family fit gave
$\Delta\approx\RelationIntercept{} \RelationSlope\,S$ with $R^{2}=\RelationRsq$
(Fig.~\ref{fig:separation-vs-gap}). However, the small panel did not support a universal rule:
leave-one-family-out slopes ranged from \RelationLooSlopeMin{} to \RelationLooSlopeMax{}, and
$R^{2}$ ranged from \RelationLooRsqMin{} to \RelationLooRsqMax{}. We therefore treated the fit
as a preliminary association. The $\omega$-conotoxin family appears in
Fig.~\ref{fig:separation-vs-gap} but not in the fit. It is not a Pfam seed family, and its
accession-based designation is not identical to the Tyr13 readout. Its plotted difference is
therefore a marker response, not a calibration error in the same sense as the five Pfam points.
Hard curation retained the marker completely in the five Pfam families and at 97.9\% in the
$\omega$-conotoxin family.

The calibration gap raised a natural comparator question: would it also appear in a model that
generates residues without SA's PCA reconstruction and argmax step?
We fitted a multiplicity-weighted profile HMM to each cleaned alignment, weighting every
designated sequence by $\rho$ and every background sequence by 1, so the HMM saw the same
split and the same multiplicity ratios as SA (Table~\ref{tab:profile-hmm-benchmark}). At
$\rho=500$, the profile HMM recovered the tracked marker at frequencies of 0.988--1.000 in the
five Pfam families, essentially reaching the target marker fraction that SA fell short of. The
contrast is consistent with a bottleneck in SA's latent-to-sequence path, but it does not isolate
that mechanism because the two models generate sequences differently. A profile HMM estimates
match-state emissions directly, whereas SA samples in PCA space and then reconstructs and
decodes a sequence. The $\omega$-conotoxin HMM produced Tyr13 in 0.834 of
its sequences and was essentially tied with SA at 0.838. Here $f_{\mathrm{eff}}=0.996$ refers
to the accession-designated mixture share, not a Tyr13 target. Both methods landed near
the marker prevalence of the designated input itself: only 19 of the 23 designated conotoxin
sequences carry Tyr13, so a generator that reproduces the designated marginals would yield the
marker in $19/23 = 0.83$ of the output. That prevalence is a reference point rather than an
upper bound, because a decoder that sharpens toward the mode can exceed it, as hard curation
does below. Closer marker matching did not imply lower language-model scores. In the matched
Kunitz $\rho=500$ comparison, ESM2 assigned pseudo-perplexity
$\SARhoFiveHundredPPL\pm\SARhoFiveHundredPPLSD$ to SA and
$\HMMRhoFiveHundredPPL\pm\HMMRhoFiveHundredPPLSD$ to the weighted profile HMM, where a lower
value means the model finds a sequence more plausible. At the same multiplicity ratio, the HMM
came closer to the target marker fraction, while SA had lower ESM2 pseudo-perplexity.

\begin{figure}[p]
  \centering
  \includegraphics[width=0.72\textwidth]{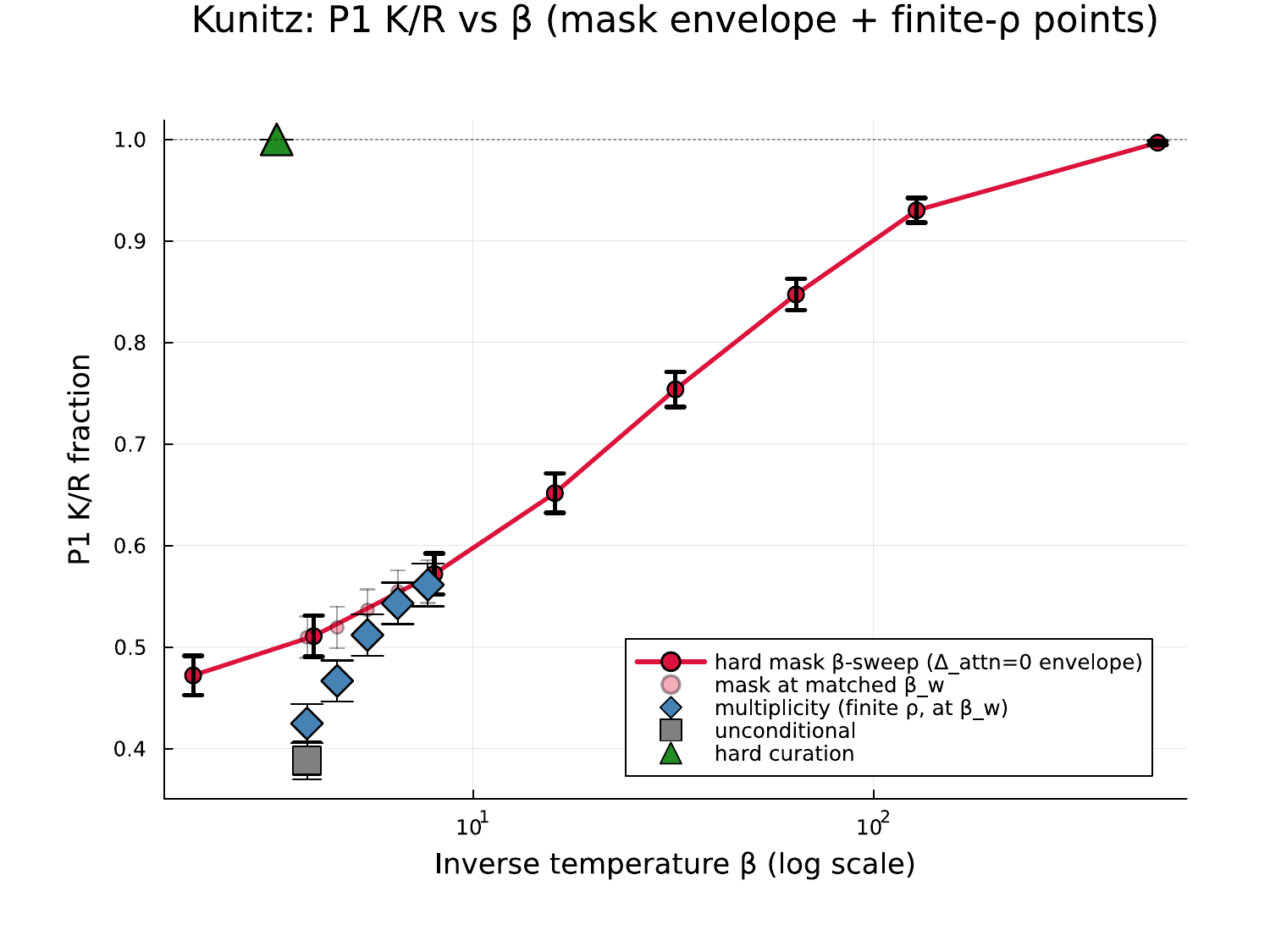}
  \caption{\textbf{Recovery depends on retrieval sharpness and conditioning strength.}
  P1~K/R fraction for the Kunitz designated subset versus inverse temperature
  $\beta$ (log scale) under the hard attention mask ($b = -\infty$ on background
  patterns; red curve, 30 chains per point, error bars are standard error across
  chains). Under the mask, all attention is restricted to designated memories. Recovery
  climbs monotonically from 0.47 at $\beta = 2$ to 1.00 at $\beta = 512$ with no plateau.
  The finite-$\rho$ multiplicity conditions (blue diamonds, each at its own
  $\beta^{*}(\vr)$) sit below this envelope at matched $\beta$: hard masking recovers more
  marker signal than soft weighting by 0.085 (SE 0.028) at $f_{\mathrm{eff}} = 0.5$,
  contracting into the noise as $\rho \to \infty$ (0.003, SE 0.030, at
  $f_{\mathrm{eff}} = 0.99$); faint red points show the mask run at each matched
  $\beta$. Because each condition's attention already matches its own target
  ($\Delta_{\mathrm{attn}} \approx 0$), this shrinking margin tracks the difference in
  intended designated mass between the two conditions rather than a residual attention
  error. Hard curation (green triangle) reaches complete transfer at low $\beta$ on
  its subset-only basis; unconditional generation (gray square) is shown for
  reference.}
  \label{fig:mask-betasweep}
\end{figure}

We next asked whether the marker signal lost between latent weighting and sequence output could
be recovered by sampling more sharply, without changing the memory
(Fig.~\ref{fig:mask-betasweep}). Two separate controls act
here: $\rho$ sets the target marker fraction, whereas the inverse temperature $\beta$ sets how
sharply the sampler retrieves individual stored patterns rather than blending them. We report
$\beta$ as the ratio $\beta/\beta^{*}$, since the crossover $\beta^{*}$ at which blending gives
way to retrieval itself shifts with $\rho$, so the ratio measures sharpness relative to each
condition's own operating point. Increasing $\beta/\beta^{*}$ from 0.5 to 3.0 at
$\rho=10$, 50, and 200 raised the decoded Kunitz P1 marker fraction but reduced diversity. At
$\rho=200$, for example, P1~K/R rose from 0.56 at $\beta^{*}$ to 0.70 at $3\beta^{*}$ while
diversity fell from 0.52 to 0.46 (Supporting Information). The target at that $\rho$ was 0.99,
so tripling the sharpness recovered about a third of the shortfall and cost 0.06 in diversity.
That left two contributors to the remaining shortfall: attention on background patterns at
finite $\rho$, and the path from a latent component to a decoded sequence. We therefore gave background patterns
zero weight inside the softmax and swept $\beta$ (Table~\ref{tab:mask-recovery}). This hard
mask puts all attention on designated patterns while leaving finite-temperature latent spread,
PCA reconstruction, and argmax decoding unchanged. Any remaining shortfall is therefore
downstream of attention allocation.
Under the hard mask, P1~K/R was only 0.50 at its operating point, $\beta^{*}=3.23$,
and reached 1.00 only near $\beta/\beta^{*}\approx160$. Complete recovery therefore takes a
sampler running two orders of magnitude sharper than its own operating point. At that sharpness
the sampler approaches the stored memories and explores less broadly,
and novelty fell from 0.43 to 0.16 over the same range. \SIRef{app:matched-beta} reports a
matched-temperature comparison of the mask against multiplicity weighting that reaches the same
conclusion. Together, these experiments showed that
sharper retrieval can compensate for latent-to-sequence loss, but only by sacrificing novelty.

Having established the conditioning mechanics on Pfam domains, we asked whether hard curation
could transfer features in a distinct disulfide-rich peptide family
(Table~\ref{tab:conotoxin-pharmacophore}). $\omega$-Conotoxins block
voltage-gated calcium channels, including Cav2.2~\cite{omegaConotoxinSAR2006}; the family
includes MVIIA (ziconotide), an approved non-opioid analgesic~\cite{ziconotideReview2004}.
Prior mutagenesis found Tyr13 essential for MVIIA and GVIA activity, making
Tyr13 a useful sequence readout here~\cite{kimTyr13Essential1995}. Lys2 and other basic residues
have also been studied as channel-binding determinants~\cite{satoBasicResidues1993}. We aligned 74 SwissProt
$\omega$-conotoxins and retained 26 positions after gap filtering. Residue numbers below denote
retained alignment columns mapped onto MVIIA, so the marker is retained column 13, corresponding
to MVIIA Tyr13. Of those 74 sequences, 23 accessions
appeared on the curated list and formed the designated subset. The remaining 51 formed the
background, for which we assumed no activity label.
Importantly, accession designation and Tyr13 were related but
distinct labels: they agreed for 64 of 74 sequences (86.5\%), with 19 Tyr13-positive sequences
in the designated set and six in the background. We therefore reported accession conditioning
and Tyr13 recovery separately; an accession-level audit identified which designations had
repository activity metadata and which were supported only by the input list.
Hard curation again shifted the selected sequence statistics. From 1{,}550 generated sequences
per condition, Tyr occurred at the marker position in 48.5\% of full-family generation versus
33.8\% of its input, and in 98.5\% of designated-subset generation versus 82.6\% of its input
(Table~\ref{tab:conotoxin-pharmacophore}). Designated-subset generation also contained 20.1\%
Lys or Arg, close to 19.7\% in its input and above the 12.0\% in full-family generation.
Multiplicity weighting transferred the marker in this family as well, and produced the largest
gain in the panel. Sweeping $\rho$ from 1 to 500 raised decoded Tyr13 from $0.463 \pm 0.040$ to
$0.838 \pm 0.035$ (Supporting Information), against 0.587 for Kunitz at the same ratio. That
endpoint sits near the 0.83 prevalence of Tyr13 among the 19 of 23 designated accessions that
carry it, which again marks where a generator reproducing the designated marginals would land
rather than a bound on what decoding can reach. The $\omega$-conotoxin split also has the
cleanest PCA separation in the panel
($S = \ConotoxinSeparation$, against \WWSeparation{}--\HomeoboxSeparation{} for the Pfam
families), qualitatively consistent with the preliminary association reported above; because
this family was held out of that fit, the agreement does not establish a predictive
relationship.

\begin{figure}[p]
  \centering
  \includegraphics[width=\textwidth]{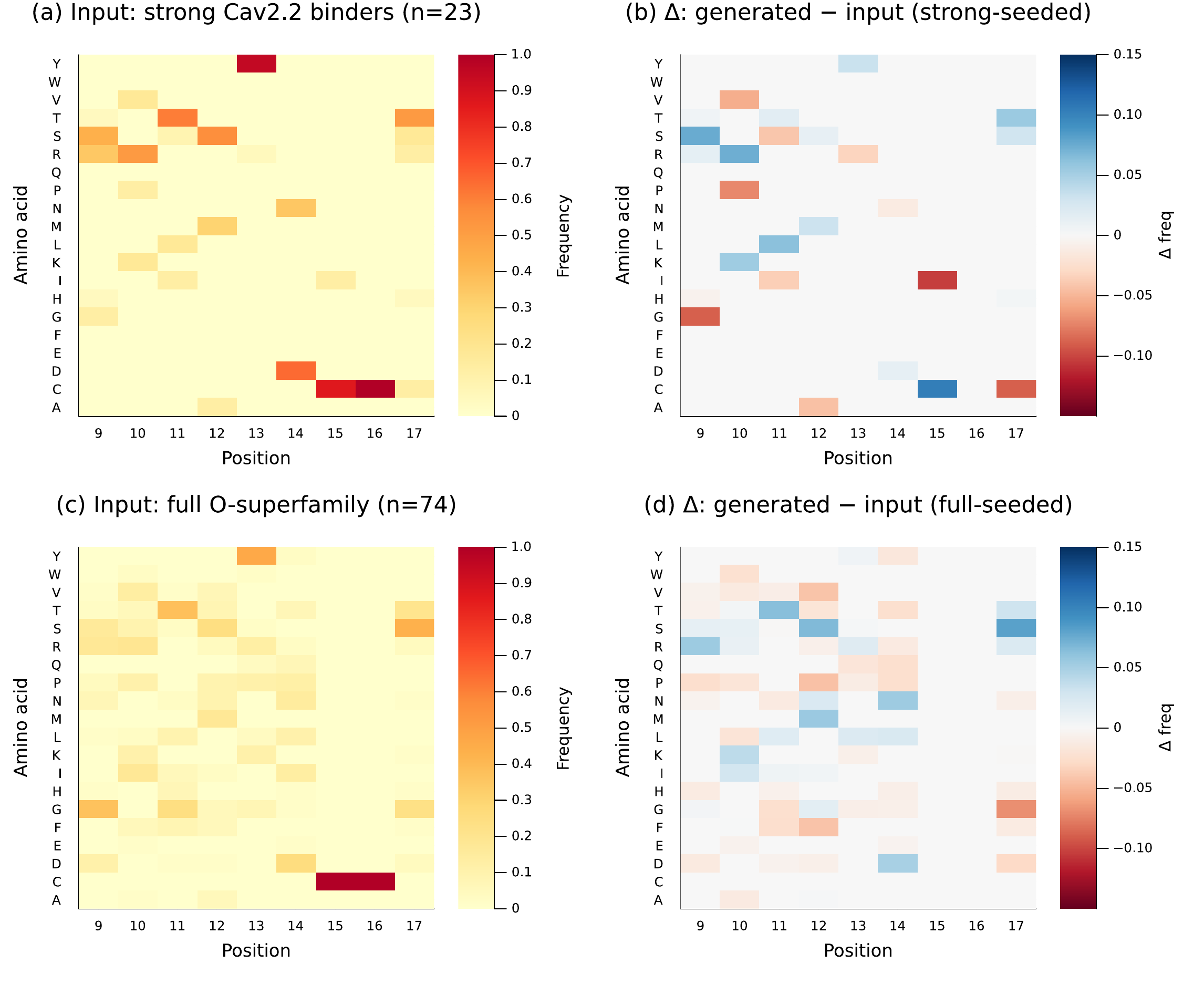}
  \caption{\textbf{Binding-loop amino acid frequencies and generation residuals for
  the $\omega$-conotoxin family (positions 9--17).}
  \textit{Left column:} Per-position amino acid frequency heatmaps for the input sequences
  (designated subset, $n=23$, panel~a; full O-superfamily, $n=74$, panel~c).
  \textit{Right column:} Residual heatmaps (generated $-$ input) showing the change in
  frequency at each position-residue pair after SA generation. Panel~(b): designated-seeded
  residual frequencies differ from the input by at most 0.16 in absolute value at every
  position-residue pair. The Tyr marker at position~13, the adjacent cysteines at
  positions~15 and~16, and the variable-loop diversity (positions~9--12) are reproduced
  within that bound. Panel~(d):
  full-seeded residuals are slightly larger, reflecting the broader sequence diversity of
  the input. Blue: enrichment; red: depletion; white: no change.}
  \label{fig:conotoxin-loop}
\end{figure}

\begin{figure}[p]
  \centering
  \includegraphics[width=\textwidth]{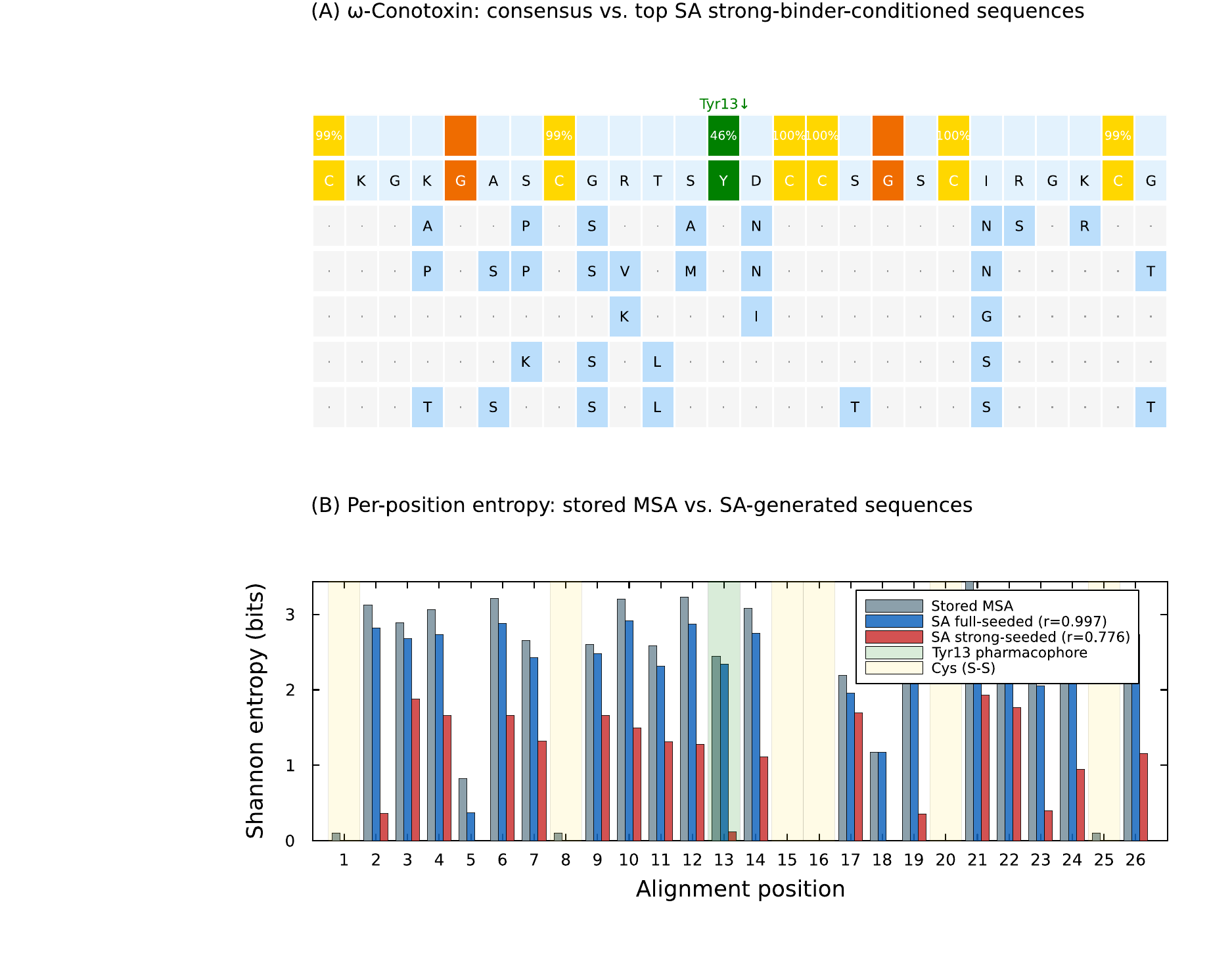}
  \caption{\textbf{Sequence-level analysis of SA-generated $\omega$-conotoxin sequences
  conditioned on the designated subset.}
  \textbf{(A)}~Alignment of the family consensus with the five highest-confidence sequences
  among the 50 modeled SA designated-subset-conditioned sequences (ranked by ESMFold pLDDT). Positions are
  colored by conservation in the stored MSA: highly conserved ($>90\%$, red),
  conserved ($70$--$90\%$, orange), moderate ($50$--$70\%$, yellow), and variable
  ($<50\%$, blue). Cysteines forming the three disulfide bonds are highlighted in
  gold; the Tyr13 marker position is highlighted in green. All five sequences
  preserve all six highly conserved cysteine positions ($6/6$) and
  carry Tyr at the marker position, while introducing $3$--$9$ substitutions
  concentrated at variable sites.
  \textbf{(B)}~Per-position Shannon entropy for the stored MSA (gray), SA full-seeded
  generation (blue), and SA designated-seeded generation (red). Full-seeded generation
  tracks the stored entropy profile closely. The designated-seeded profile tracks it
  less closely, reflecting lower diversity after conditioning on the 23-sequence
  designated subset, including at Tyr13 and co-varying loop residues.}
  \label{fig:conotoxin-sequence-analysis}
\end{figure}

\begin{figure}[p]
  \centering
  \includegraphics[width=\textwidth]{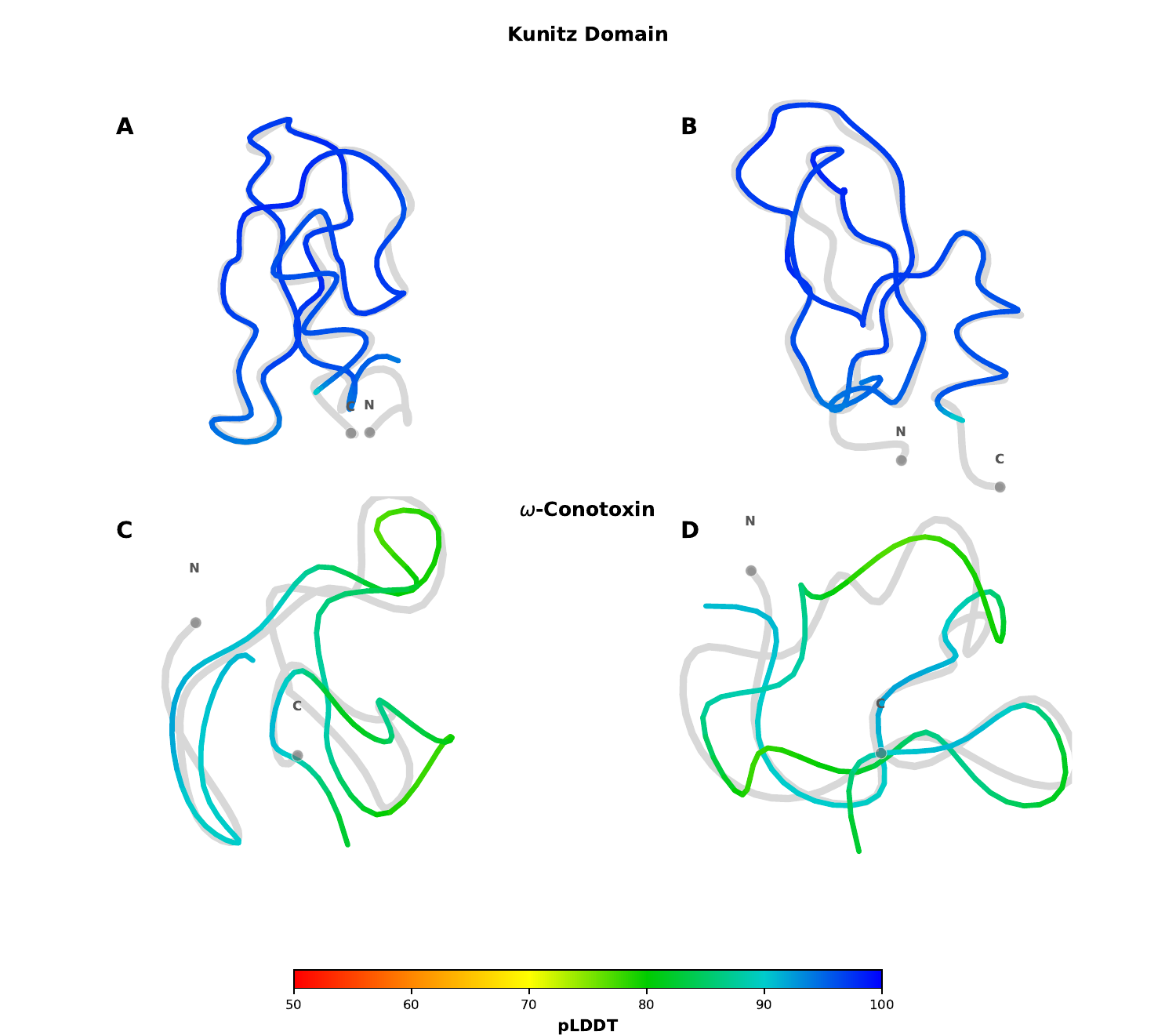}
  \caption{\textbf{Structure models place selected SA-generated sequences near
  experimental family references.}
  For each family, the single highest TM-score SA designated-subset-conditioned sequence
  (colored by per-residue pLDDT confidence) is superimposed onto the experimental
  reference structure (gray, semi-transparent) and shown in two orthogonal views
  (front and $90^{\circ}$ rotation) to reveal the full three-dimensional architecture.
  \textbf{(A)}~Kunitz domain, front view. \textbf{(B)}~Kunitz domain, $90^{\circ}$
  rotation. The SA-generated variant aligns to the BPTI crystal structure
  (1BPI chain~A) with TM $= 0.86$, reproducing the double-loop architecture,
  three-stranded antiparallel $\beta$-sheet, and $\alpha$-helical turn;
  mean C$\alpha$ confidence is 96.3.
  \textbf{(C)}~$\omega$-Conotoxin, front view. \textbf{(D)}~$\omega$-Conotoxin,
  $90^{\circ}$ rotation. The SA-generated variant aligns to the MVIIA NMR structure
  (1OMG chain~A) with TM $= 0.53$; its predicted backbone follows the reference
  in this illustrative alignment, with mean C$\alpha$ confidence 85.7.
  These highest-scoring examples are not group-level validation; group summaries
  appear in Table~\ref{tab:structure-model-comparison}.
  Colorbar: pLDDT confidence scale (red $= 50$, blue $= 100$).
  Structures predicted by ESMFold~\cite{esmfold2022}; alignment by
  TM-align~\cite{tmAlign2005}.}
  \label{fig:fold-superposition}
\end{figure}

The broader sequence pattern was retained alongside these marker shifts. Frequency heatmaps
preserved Tyr13, the cysteine framework, and variable-loop diversity
(Fig.~\ref{fig:conotoxin-loop}); measured against the full-family input entropy profile,
per-position correlations were $r=0.776$ for designated-subset generation and $r=0.997$ for
full-family generation. Among the 50 modeled designated-subset sequences, the five
highest-pLDDT sequences retained all six highly conserved cysteine positions while introducing
3--9 substitutions at variable sites (Fig.~\ref{fig:conotoxin-sequence-analysis}).
Comparison with 12 published structure-activity relationship positions gave a consistent but
strictly sequence-level result. Designated-subset generation retained Tyr13 at 98.5\% and all
six invariant disulfide-forming cysteines. It also increased K/R at reported SAR
positions, including the Arg10-aligned column (83.4\% versus 69.6\% in the designated input) and
position~21 (52.4\% versus 47.8\%). These shifts describe residue frequencies; they do not test
mutation effects in the generated backgrounds~\cite{kimTyr13Essential1995,satoBasicResidues1993,lewisConotoxinSAR2012}.
The single highest-TM-score modeled sequence also aligned visually with the short experimental
reference backbone (Fig.~\ref{fig:fold-superposition}). Thus, without
retraining or a Cav2.2 target structure, hard curation transferred the designated accessions'
sequence statistics into a new library; it did not establish target binding.

Finally, we asked whether marker control came at the expense of global sequence or structural
plausibility (\SIFigRef{fig:plddt-tmscore-scatter}). We scored 50 Kunitz sequences from each of
five sources: stored, full-family SA, marker-positive SA, marker-negative SA, and unconditional
profile-HMM emission. Across those sources, ESMFold assigned similar model scores to SA and
stored sequences (Table~\ref{tab:structure-validation}). Mean pLDDT was $90.4\pm1.2$ for
full-family SA, $90.7\pm1.1$ for marker-positive SA, $91.0\pm1.2$ for marker-negative SA, and
$89.3\pm3.2$ for stored sequences, with every SA sequence above 70. Corresponding TM-scores
against the experimental Kunitz structure (1BPI, chain~A) were $0.84\pm0.01$, $0.83\pm0.01$,
$0.85\pm0.01$, and $0.83\pm0.02$. Sequences emitted by the unconditional profile HMM, built with
default \texttt{hmmbuild} settings on the seed alignment rather than the weighted build used for
the matched benchmark, scored lower and more variably, with mean pLDDT $60.4\pm13.1$, 20\% above
70, and mean TM-score $0.60\pm0.19$. AlphaFold2-ptm provided a second structure predictor and
gave the same Kunitz pattern, with TM-scores of
0.83--0.85 at pLDDT 93--95 (Table~\ref{tab:structure-model-comparison})
~\cite{jumperHighlyAccurate2021,mirdita2022colabfold}. The $\omega$-conotoxins, at 24 to 35
residues, scored lower against their reference structure (1OMG, chain~A), but so did the stored
natural sequences scored alongside them. Under AlphaFold2-ptm, TM-scores were $0.35\pm0.11$ for stored
$\omega$-conotoxins, $0.35\pm0.13$ for full-family generation, and $0.47\pm0.02$ for
designated-subset generation, at pLDDT 68--80; ESMFold gave $0.43\pm0.07$, $0.47\pm0.03$, and
$0.47\pm0.03$ for the same three groups (Table~\ref{tab:structure-model-comparison}). The
designated-subset group had a higher mean TM-score than the stored group under both predictors.
The full-family group also had a higher mean under ESMFold; under AlphaFold2-ptm, its mean exceeded
the stored mean by only 0.0027, negligible relative to the within-group standard deviations.
Because every score was normalized to the 25-residue reference chain and the stored and generated
distributions overlapped, neither predictor showed a lower generated-group mean relative to the
stored comparison.
The low absolute values do not by themselves distinguish short-chain TM-score behavior from
structural differences among the peptides.

Sequence scoring with ESM2 also placed SA near the stored family distribution
(Table~\ref{tab:structure-validation}). Pseudo-perplexity was $4.78\pm0.64$ for full-family SA,
$4.72\pm0.60$ for marker-positive SA, and $4.97\pm0.98$ for stored sequences, compared with
$16.1\pm4.2$ for unconditional HMM emission. The weighted profile HMM of the matched benchmark,
fitted to the cleaned alignment without entropy weighting, instead scored
$\HMMRhoFiveHundredPPL\pm\HMMRhoFiveHundredPPLSD$, so the contrast drawn here is with
unconditional emission and not with profile HMMs in general.
Sequence-marker metrics from the same unconditional baseline confirmed the distinction among
generation, conditioning, and resampling. Across five HMMER3 emissions and five bootstrap
resamples~\cite{hmmer3}, full-family SA gave a P1~K/R fraction of $0.38\pm0.03$ and KL divergence
$0.0076\pm0.0009$, compared with $0.17\pm0.04$ and $0.027\pm0.002$ for HMM emission.
Bootstrap resampling matched composition most closely (KL $\approx0.001$) but produced no novel
sequences. Marker-positive SA, by contrast, reached P1~K/R $1.0\pm0.0$, a conditioning control
that neither unconditional baseline provided (Table~\ref{tab:structure-validation}). These
model-based evaluations did not demonstrate function. They indicate that marker control remained
compatible with family-level sequence plausibility and Kunitz-like predicted folds.

\section{Discussion}
Pattern multiplicity in the modern Hopfield energy provides a training-free way to condition
protein sequence generation on user-designated subsets, since the multiplicity weights add a
logit bias to the Boltzmann distribution whose corresponding continuous-time equilibrium
target is an exact Gaussian mixture. Finite-step ULA retains discretization and mixing error,
but its decoded marker fractions and amino acid compositions were similar to independent
exact-equilibrium draws, and exact sampling reproduced the large gap between
$f_{\mathrm{eff}}=0.996$ and decoded P1~K/R recovery at $\rho=500$, indicating that
finite-step ULA error was not the main source of that gap. A single scalar $\rho$ continuously
interpolates between unconditioned generation and hard subset curation, and the value required
for a target designated fraction follows directly from the family composition
(Eq.~\ref{eq:rho-inverse}). In the Kunitz domain, curated memory retained the selected
K/R-at-P1 marker with as few as three designated inputs. Full-family multiplicity weighting
increased decoded marker frequency from $0.382\pm0.031$ at $\rho=1$ to
\KunitzFobsFiveHundred{} $\pm$ \KunitzFobsFiveHundredSD{} at $\rho=500$. Across five Pfam
families, PCA-space geometry was associated with conditioning effectiveness.

The matched profile-HMM benchmark narrows the claim of advantage. When the objective is direct
recovery of a selected single-position marginal, sequence-weighted profile HMMs were stronger:
at $\rho=500$ they reproduced the tracked marker at 0.988--1.000 across the five Pfam
families, whereas SA retained a latent-to-sequence calibration gap. Conversely, in the matched
Kunitz sample, ESM2 assigned lower pseudo-perplexity to SA than to the weighted profile HMM
($\SARhoFiveHundredPPL\pm\SARhoFiveHundredPPLSD$ versus
$\HMMRhoFiveHundredPPL\pm\HMMRhoFiveHundredPPLSD$). These observations support a tradeoff,
not a general dominance claim: a profile HMM transfers position-specific weights directly,
whereas SA samples around complete stored sequences but must decode through a linear latent
representation. Neither marker recovery nor language-model scoring demonstrates biochemical
activity. Structure-conditioned inverse folding and pretrained generative models are also
important comparators, but they consume a target backbone or large external training corpora and
therefore do not constitute matched training-free baselines under the information constraints
studied here. Prospective comparisons using experimentally measured multi-position phenotypes
remain necessary.

We observed a calibration gap between latent designated mass and decoded marker frequency.
The mean attention differed from $f_{\mathrm{eff}}$ by at most \AttnMaxDevPts{} percentage
points across the canonical sweeps, so the remaining gap arises as latent states pass through
finite-temperature noise, unit-normalized PCA coordinates, affine reconstruction, and argmax
decoding (Eq.~\ref{eq:decoder-pushforward}). When the designated and background subsets
overlap in PCA space, the energy bias may not change the decoded residue identity. Across five
Pfam families, the exploratory linear fit had $R^{2}=\RelationRsq$, but it was sensitive to
individual families, since leave-one-family-out slopes ranged from \RelationLooSlopeMin{} to
\RelationLooSlopeMax{} and $R^{2}$ from \RelationLooRsqMin{} to \RelationLooRsqMax{}. The
separation index should therefore be treated as a descriptive diagnostic rather than an
\emph{a priori} predictor or decision threshold, and broader prospective validation is
required before it can guide the choice between multiplicity weighting and hard curation. The
Kunitz loss therefore arose mainly downstream of latent weighting. The same
multiplicity weights have been used to generate synthetic patients from small clinical
cohorts~\cite{varnerSAPatient2026}, where the generated values are clinical measurements that
never have to be turned back into discrete symbols. Protein generation has no such option, since
every sample must be reconstructed from PCA coordinates and then reduced to a single amino acid
at each position. Finite-temperature variation, reconstruction, and decoding compressed the
marker response.

Small characterized sets of the kind hard curation requires arise routinely from screens,
functional selections, and literature curation, and SA can expand such a set into additional
sequences that share its features. In the Kunitz experiment, however, selection and evaluation
both used K/R at P1, so complete recovery shows marker retention rather than functional
validation, generated candidates would still require experimental screening, and the method
does not assign a biological meaning to the designation. The scaling study showed that
diversity increased with the number of designated inputs and changed little beyond
$K_{\mathrm{des}}=20$. The $\omega$-conotoxin experiment provides a second example of
conditioning from a small designated subset, and it also contrasts full-family with
curated-subset seeding. In this experiment, full-family seeding favored broader diversity,
whereas curated seeding favored retention of a specific sequence marker. That contrast is sharp
in the O-superfamily,
which shares the C--C--CC--C--C cysteine framework but spans distinct channel targets and
selectivity profiles. Full-family seeding yielded less Tyr13, consistent with a PCA encoding
that reflects the broader sequence diversity. Curated seeding instead
restricts the PCA basis to variation among the 23 designated accessions, and it recovered
Tyr13 at 98.5\% frequency. Because the accession designation and the Tyr13 marker agree for only
86.5\% of stored sequences, these two outcomes are related but not interchangeable.
Hard curation can achieve high marker retention by restricting the memory to the designated
subset, even when full-family weighting has a sizable marker gap. That retention costs sequence
diversity and novelty (0.45 vs.\ 0.59), a cost that reflects the
narrower PCA basis and is consistent with the Kunitz scaling study.

Several limitations should be acknowledged. The Supporting Information AlphaFold2-multimer
diagnostic produced
low-confidence predictions for both generated and control sequences, with iPTM of
approximately 0.10 and interface pLDDT of approximately 37--41. Permutation tests detected no
group differences, but nonsignificance at sample sizes 10/10/5 does not establish equivalence
or preserved binding geometry, so those calculations provide no positive evidence about Cav2.2
interaction. The cross-family regression spans only five Pfam families, with
$\omega$-conotoxin displayed externally rather than included in the fit, and its
leave-one-family-out sensitivity and nonmonotonic points preclude a validated predictive rule,
so additional prospectively selected families are needed to characterize the
$S$--$\Delta$ relationship. Our Pfam splits are defined by data-selected single marker
positions (P1 in Kunitz, Trp in SH3, a specificity-loop residue in WW, Gln at position~50 in
Homeobox, and H/N in the Forkhead recognition helix), and multi-position phenotypes may show
different calibration behavior because they depend on coordinated residue identities. We have
also not experimentally validated the binding activity of generated sequences, and marker
fidelity is not evidence of functional binding. For the $\omega$-conotoxin family, we compared
generated residue frequencies with positions reported by alanine-scanning and site-directed
mutagenesis studies~\cite{kimTyr13Essential1995,satoBasicResidues1993,lewisConotoxinSAR2012},
and designated-seeded generation retained the cysteine scaffold and Tyr13 signal and shifted
several reported positions (Table~\ref{tab:sar-agreement}), but this does not establish their
effects in the new sequence backgrounds. For the Kunitz domain, deep mutational scanning data
for the BPTI--trypsin interaction are available~\cite{heyneBPTIDMS2021} and could provide
a quantitative binding-landscape comparison, one of several extensions that follow from these
limitations. The binary
designated-background split generalizes to multi-class conditioning by assigning different
multiplicities to different functional categories, and the PCA bottleneck that limits
conditioning effectiveness might be reduced by alternative sequence encodings, since learned
representations from protein language models may separate functional subsets better than
linear PCA does. The $\beta$ lever suggests that adaptive temperature schedules, increasing
$\beta$ during sampling to sharpen the multiplicity bias, could improve marker transfer
without permanently sacrificing diversity. All of these extensions remain within the analytic,
training-free framework that makes stochastic attention practical for the scarce-data regime.

\section{Materials and Methods}
\subsection{From Hopfield Energy to Multiplicity-Weighted Score Function}\label{sec:theory}

Let $\vX = [\vm_1, \ldots, \vm_K] \in \R^{d \times K}$ be a memory matrix whose columns are
unit-norm, PCA-encoded protein sequences from a family alignment of $K$ members, each represented
in $d$ dimensions. The modern Hopfield energy~\cite{ramsauerHopfieldNetworksAll2021} at state
$\vxi \in \R^d$ is given by:
\[
E(\vxi) = \tfrac{1}{2}\|\vxi\|^{2}
- \tfrac{1}{\beta}\log\sum_{k}\exp(\beta\,\vm_{k}^{\!\top}\vxi),
\]
where $\beta > 0$ is an inverse temperature. Its score function,
\[
\nabla_{\vxi}\log p_{\beta}(\vxi)
= \beta[\vX\,\softmax(\beta\,\vX^{\!\top}\vxi) - \vxi],
\]
is exact and requires one attention operation. Applying the unadjusted Langevin algorithm
(ULA) with score step size $\alpha/\beta$, where $\alpha \in (0,1)$, gives the stochastic
attention update~\cite{varnerSAProtein2026}:
\begin{equation}\label{eq:ula}
  \vxi_{t+1}
    = (1-\alpha)\,\vxi_{t}
    + \alpha\,\vX\,\softmax\!\bigl(\beta\,\vX^{\!\top}\vxi_{t}\bigr)
    + \sqrt{2\alpha/\beta}\;\veps_{t},
  \qquad \veps_{t} \sim \mathcal{N}(\mathbf{0},\,\mathbf{I}_{d}).
\end{equation}
We extend this framework to condition generation on a designated subset by assigning a
multiplicity weight $r_k > 0$ to each stored pattern. The \emph{multiplicity-weighted} Hopfield
energy is given by:
\begin{equation}\label{eq:weighted-energy}
  \boxed{
  E_{\vr}(\vxi) = \tfrac{1}{2}\|\vxi\|^{2}
    - \tfrac{1}{\beta}\log\sum_{k=1}^{K} r_k\,\exp\!\bigl(\beta\,\vm_{k}^{\!\top}\vxi\bigr),}
\end{equation}
where $\vr = (r_1, \ldots, r_K)^{\!\top}$.
When $r_k = 1$ for all $k$, Eq.~\eqref{eq:weighted-energy} reduces to the standard energy, and
patterns with higher multiplicity deepen the energy wells in their vicinity. For integer $r_k$
this is exactly equivalent to storing $r_k$ copies of pattern $\vm_k$ without the
$\mathcal{O}(d \cdot \sum_k r_k)$ memory cost, and for positive real $r_k$ it is the continuous
generalization of that replication.

\begin{samepage}
\begin{proposition}[Exact Gaussian-mixture target]\label{prop:gmm-target}
For unit-norm memories, the Gibbs density associated with
Eq.~\eqref{eq:weighted-energy}, and hence the stationary law of the corresponding
continuous-time overdamped Langevin diffusion, is given by:
\begin{equation}\label{eq:gmm-target}
  \boxed{
  p_{\vr}(\vxi)
  = \sum_{k=1}^{K} w_k\,
    \mathcal{N}\!\left(\vxi;\vm_k,\beta^{-1}\mathbf{I}_d\right),
  \qquad
  w_k = \frac{r_k}{\sum_j r_j}.}
\end{equation}
Independent equilibrium draws therefore require only
$k\sim\operatorname{Categorical}(\mathbf{w})$ followed by
$\vxi=\vm_k+\beta^{-1/2}\veps$, with
$\veps\sim\mathcal{N}(\mathbf{0},\mathbf{I}_d)$.
\end{proposition}
\end{samepage}

The derivation is given in the Supporting Information.
For unit-norm memories we can therefore draw equilibrium samples directly from the
mixture, and we use those draws as the reference in the exact-equilibrium benchmark.
We still run ULA for the reported generation runs, not because the target requires
it, but to stay comparable with the original stochastic-attention workflow and to
measure what finite steps cost.
Equation~\eqref{eq:gmm-target} is exact for the idealized dynamics that
move smoothly in time. The updates we actually run, Eqs.~\eqref{eq:ula} and
\eqref{eq:weighted-ula} below, advance in finite jumps of size $\alpha$, and a
stepped chain does not in general settle on the same distribution as the smooth one.
Two errors therefore remain at nonzero $\alpha$, a step-size bias that persists no
matter how long we run, and the error from stopping after finitely many
steps~\cite{durmusMoulinesULA2017}.

\begin{proposition}[Score function]\label{prop:score}
The score function of $p_{\vr}(\vxi) \propto \exp(-\beta\,E_{\vr}(\vxi))$ is given by:
\[
\nabla_{\vxi}\log p_{\vr}(\vxi)
= \beta[\vX\,\softmax(\beta\,\vX^{\!\top}\vxi + \log\vr) - \vxi],
\]
where $\log\vr = (\log r_1, \ldots, \log r_K)^{\!\top}$ is the elementwise log-multiplicity.
\end{proposition}

The proof follows from the identity $r_k\,\exp(a_k) = \exp(a_k + \log r_k)$: the weights are
absorbed into the softmax logits. The resulting Langevin update is given by:
\begin{equation}\label{eq:weighted-ula}
  \boxed{
  \vxi_{t+1}
    = (1-\alpha)\,\vxi_{t}
    + \alpha\,\vX\,\softmax\!\bigl(\beta\,\vX^{\!\top}\vxi_{t} + \log\vr\bigr)
    + \sqrt{2\alpha/\beta}\;\veps_{t}.}
\end{equation}
The \emph{only} change from Eq.~\eqref{eq:ula} is the addition of $\log r_k$ to each logit
before the softmax. The memory matrix $\vX$ is unchanged, the noise schedule is unchanged, and
the sampler retains its $\mathcal{O}(dK)$ per-step cost.

\subsection{Multiplicity Ratio and the Entropy Crossover}\label{sec:phase}

The user designates $K_{\mathrm{des}}$ of the $K$ stored patterns, leaving the
remaining $K_{\mathrm{bg}} = K - K_{\mathrm{des}}$ as background. Each stored pattern is one
sequence of the alignment, so $K_{\mathrm{des}}$ counts sequences and not alignment positions.
In the Kunitz family studied here, the memory holds $K = 99$ aligned domains, of which the
$K_{\mathrm{des}} = 32$ carrying Lys or Arg at the P1 contact position are designated and the
remaining $K_{\mathrm{bg}} = 67$ are background. The designation is
an input to the method rather than something the method infers, and nothing below
depends on what it means. A list of accession IDs would serve as well as a marker residue,
because the sampler sees only the resulting labels. We give every designated pattern the same
multiplicity and every background pattern the same multiplicity, and we write $\rho$
for the ratio between the two. Fixing the background multiplicity at $1$ sets the
scale, so $r_k = \rho$ for designated patterns and $r_k = 1$ for background
patterns, and $\rho$ is then the only free parameter.

The designated patterns therefore carry total multiplicity $K_{\mathrm{des}}\rho$
out of $K_{\mathrm{des}}\rho + K_{\mathrm{bg}}$ in all, so by
Eq.~\eqref{eq:gmm-target} the probability that a draw comes from a designated
component is given by:
\begin{equation}\label{eq:f-eff}
  f_{\mathrm{eff}}(\rho)
  = \frac{K_{\mathrm{des}}\,\rho}{K_{\mathrm{des}}\,\rho + K_{\mathrm{bg}}} ,
\end{equation}
which we call the effective designated fraction. At $\rho=1$ it equals
$K_{\mathrm{des}}/K$, the designated share of the family, and as $\rho \to \infty$
it approaches $1$. Inverting Eq.~\eqref{eq:f-eff} gives the multiplicity ratio
needed for a target fraction $f$:
\begin{equation}\label{eq:rho-inverse}
  \rho(f) = \frac{f \cdot K_{\mathrm{bg}}}{K_{\mathrm{des}}\,(1-f)}.
\end{equation}
Choosing $\rho$ fixes the mixture we are aiming at, but it says nothing about how
sharply the sampler retrieves from that mixture. That is the job of $\beta$, and
$\beta$ has to be chosen before we can run anything. The division of labor is clean in
one direction: for unit-norm memories $\beta$ sets the width $\beta^{-1}$ shared by
every mixture component, so it controls how much neighboring components overlap, which
component a given state most likely came from, and therefore what the decoder returns,
but it does not touch the mixture weights $w_k$ or $f_{\mathrm{eff}}$. It is not clean
in the other direction, and the rest of this subsection is about how $\rho$ moves the
$\beta$ we should use.

The choice matters because both extremes of $\beta$ are useless. The attention weights
in Eq.~\eqref{eq:weighted-ula} are $\softmax(\beta\vX^{\!\top}\vxi + \log\vr)$. When
$\beta$ is small the logits are nearly flat, every stored pattern draws close to its
multiplicity weight $w_k$, and the sampler steps toward a weighted average of the whole
family, which decodes to a consensus-like blur. When $\beta$ is large the softmax
collapses onto whichever pattern the state is closest to, and the sampler sits on that
stored sequence and reproduces it. Useful generation happens between these two
regimes, and the attention entropy is what tells us where the transition is.
Following the base SA method~\cite{varnerSAProtein2026}, we track the Shannon entropy
$H_{\vr}(\beta)$ of those weights as $\beta$ increases, and we operate at the
onset of its drop. We call that drop the entropy crossover, and we locate its onset at
the point of maximum downward curvature of $H_{\vr}$ against $\log\beta$, writing the
inverse temperature there as $\beta^{*}$ (Fig.~\ref{fig:phase-transition}). This is a
convention for picking a reproducible operating point, not a claim that anything
thermodynamic happens there.

For standard SA the operating point does not have to be found by sweeping at all. The
companion paper~\cite{varnerSAProtein2026} establishes, from a concentration-of-measure
argument on the attention entropy, that the unweighted sampler ($\rho=1$) satisfies:
\begin{equation}\label{eq:beta-star}
  \beta^{*} \;\approx\; 1.57 + 0.28\sqrt{d},
\end{equation}
with $R^{2}=0.97$ across eight Pfam families spanning $d \in [18, 186]$, and nearly
independent of $K$ for $K \geq 30$. The alignment alone therefore fixes $\beta^{*}$.
Multiplicity weighting breaks that shortcut. The biases $\log r_k$ sit in the logits before $\beta$ does
anything, so at $\beta=0$ the attention is already uneven: its weights are
$w_k = r_k/\sum_j r_j$ and its entropy is $H_{\vr}(0) = -\sum_k w_k \log w_k$, below
the uniform value $\log K$ that standard SA starts from. Every entropy curve therefore
starts lower and its drop moves to larger $\beta$
(Fig.~\ref{fig:phase-transition}), so for $\rho > 1$ we abandon
Eq.~\eqref{eq:beta-star} and re-find $\beta^{*}(\rho)$ by sweeping $\beta$ at each
$\rho$. Every experiment below runs at the $\beta^{*}(\rho)$ found this way. On the
Kunitz family it rose monotonically from 4.4 at $\rho=1$ to 9.3 at $\rho=1{,}000$, and
over that same range the effective pattern count
$K_{\mathrm{eff}}(\vr) = (\sum_k r_k)^{2}/\sum_k r_k^{2}$ fell from 99 to 32.1. That
count equals $K$ when the multiplicities are all equal and falls toward
$K_{\mathrm{des}}$ as $\rho\to\infty$, so it summarizes how concentrated the weights
have become, and we report it beside the $\beta^{*}$ shift as a companion observation
rather than derive either from the other. The two are not the same
quantity, since $\log K_{\mathrm{eff}}$ is the Renyi-2 entropy of the weights and
equals the Shannon entropy $H_{\vr}(0)$ only when the weights are equal.

\subsection{The Calibration Gap}\label{sec:gap}

Setting $\rho$ fixes $f_{\mathrm{eff}}$, the share of the equilibrium mixture assigned to
designated components. At exact equilibrium, attention is the posterior responsibility of each
mixture component, so its expected designated mass is also $f_{\mathrm{eff}}$. A finite ULA run
only estimates that expectation, which is why we measure $\bar{a}_{\mathrm{des}}$ directly
(Eq.~\ref{eq:a-des}).

The discrete output is a different quantity. We write $f_{\mathrm{obs}}$ for the fraction of
decoded sequences carrying the tracked marker and define
$\Delta=f_{\mathrm{eff}}-f_{\mathrm{obs}}$. When designation and marker are the same label, as
in the five Pfam experiments, $\Delta$ is a calibration gap across latent sampling, PCA
reconstruction, and argmax decoding. It can be negative because decoding may produce more
marker-positive sequences than the designated mixture share. For $\omega$-conotoxin,
designation is an accession list and Tyr13 is a separate readout, so the same difference is
descriptive rather than a calibration error.

Because the target is exactly a mixture, we can describe the complete reconstruction and
decoding path by conditioning on which component produced the sample.
Let $C$ be that component and let $\mathcal{D}$ be the index set of the designated patterns, so
that $C\in\mathcal{D}$ says the draw came from a designated component. We define:
\[
\begin{aligned}
q_{\mathrm{des}}
  &= \Pr(\text{decoded marker positive}\mid C\in\mathcal{D}),\\
q_{\mathrm{bg}}
  &= \Pr(\text{decoded marker positive}\mid C\notin\mathcal{D}).
\end{aligned}
\]
These are group-specific latent-to-sequence transfer rates. Each includes
finite-temperature spread around a component as well as reconstruction and decoding.
The rate $q_{\mathrm{des}}$ is how often a draw from a designated component becomes a
marker-positive sequence; $q_{\mathrm{bg}}$ is the corresponding rate for a background draw.
Combining these rates gives:
\begin{equation}\label{eq:decoder-pushforward}
  f_{\mathrm{obs}}
  = f_{\mathrm{eff}}q_{\mathrm{des}}
  + (1-f_{\mathrm{eff}})q_{\mathrm{bg}},
  \qquad
  \Delta
  = f_{\mathrm{eff}}(1-q_{\mathrm{des}})
  - (1-f_{\mathrm{eff}})q_{\mathrm{bg}}.
\end{equation}
Thus $f_{\mathrm{obs}}=f_{\mathrm{eff}}$ is not something the algebra requires at
$\rho=1$. It holds only when the decoder's false negatives happen to balance its
false positives. The calibration gap is therefore a quantity to be measured across
the whole path from latent state to sequence, not a diagnostic for tuning that
path. Matching $f_{\mathrm{obs}}$ to $f_{\mathrm{eff}}$ in particular does not
identify a correct PCA scaling, because a decoder that carries no information at
all can match the two by trading one error rate against the other.
The transfer rates depend on how the designated and background sequences sit relative to each
other in PCA space, so we need one number for that geometry, comparable across families.
We define the separation index
$S = (\bar{c}_{\mathrm{within}} - \bar{c}_{\mathrm{between}}) /
[\tfrac{1}{2}(\sigma_{\mathrm{within}} + \sigma_{\mathrm{between}})]$, where $\bar{c}$ and
$\sigma$ denote the mean and standard deviation of pairwise cosine similarities within and
between designated and background groups in PCA space. This is a Fisher-type ratio defined for this study
on pairwise cosine similarities, not the classical Fisher discriminant criterion,
which is computed from class means and scatter matrices.

\subsection{Data Curation and Sequence Encoding}

For each experiment, we followed the stochastic attention pipeline
of~\cite{varnerSAProtein2026} without modification to the core sampler. Seed alignments were
downloaded from InterPro/Pfam in Stockholm format, while the $\omega$-conotoxin O-superfamily
was compiled from SwissProt and aligned with MAFFT~\cite{mafft2013}. We removed columns with
$>50\%$ gaps and sequences with $>30\%$ gaps in the remaining columns, raising the sequence
threshold to $>40\%$ for the $\omega$-conotoxins to accommodate their broader length range, then one-hot encoded the
cleaned alignment with 20 amino acid channels per position and mapped gaps to zero vectors.
Principal component analysis retained 95\% of the variance, and the resulting columns were
normalized to unit norm.

Each ULA chain used step size $\alpha=0.01$ and ran for $T=5{,}000$ steps, starting near a
stored pattern with patterns assigned cyclically across the memory and with independent local
random number generators (RNGs) generating the initial perturbations and Langevin noise. Samples were collected after
2{,}000 burn-in steps and then every 100 steps. We tested five Pfam families, Kunitz
(PF00014, $K=99$, $L=53$), SH3 (PF00018, $K=55$, $L=48$), WW (PF00397, $K=420$, $L=31$),
Homeobox (PF00046, $K=136$, $L=57$), and Forkhead (PF00250, $K=246$, $L=87$), together with
the $\omega$-conotoxin O-superfamily ($K=74$, $L=26$), which was curated from
SwissProt. For each Pfam family, we defined a binary split using a data-selected
single-position sequence marker, with the residue class motivated by biological literature
where available but not treated as a direct activity label. The marker was the column with the
highest Lys frequency for Kunitz, denoted P1, whose designated subset contained the sequences
with K or R at that position ($K_{\mathrm{des}}=32$); the peptide-groove column with the
highest Trp frequency among
columns containing Trp in 15--85\% of sequences for SH3 ($K_{\mathrm{des}}=33$); the
highest-entropy column in the middle third of the alignment for WW ($K_{\mathrm{des}}=69$);
Gln at position~50 in the recognition helix for Homeobox ($K_{\mathrm{des}}=102$); and His or
Asn at the H3 recognition-helix position for Forkhead ($K_{\mathrm{des}}=122$). For
$\omega$-conotoxin, the designation was instead defined by 23 accession IDs in the study's
curated input FASTA, with the other 51 accessions serving as conditioning background; we did
not treat the background as experimentally inactive. The tracked accession audit records the
designation, Tyr13
marker, and available repository activity metadata, and only 5 designated and 2 background
accessions have exact accession matches in that activity table, while the accession designation
and Tyr13 marker agree for 64 of 74 sequences (86.5\%).

\subsection{SA Sampling and Evaluation}

For multiplicity-weighted runs, the multiplicity vector was set to $r_k = \rho$ for designated
patterns and $r_k = 1$ for background patterns, and the operating inverse temperature
$\beta^{*}(\vr)$ was selected as the entropy-crossover onset, which is the point of maximum
downward curvature of weighted attention entropy with respect to $\log\beta$. We evaluated
entropy on a fixed logarithmic grid over $\beta\in[0.1,500]$ and averaged it over all stored
memories, so the resulting operating point does not depend on alignment order. The grid used 50 points
for the canonical family sweeps and the hard-mask comparison, 60 points for the exact-equilibrium
benchmark, and 80 points for the entropy curves in Fig.~\ref{fig:phase-transition}. This is a
descriptive operating point rather than an estimate of a thermodynamic phase transition, and
entropy is evaluated at stored memory vectors rather than at samples. Alongside $\beta^{*}$ we
report the effective pattern count
\begin{equation}\label{eq:k-eff}
  K_{\mathrm{eff}}(\vr) = \frac{\bigl(\sum_k r_k\bigr)^{2}}{\sum_k r_k^{2}},
\end{equation}
which measures how concentrated the multiplicity weights have become. It equals $K$ when the
multiplicities are all equal and falls toward $K_{\mathrm{des}}$ as $\rho\rightarrow\infty$. Twenty to fifty independently seeded chains were run per
condition, yielding 620--1{,}550 retained states after burn-in and thinning, with the canonical
six-family sweeps using five replicate runs of 20 chains and reported uncertainty computed
across replicate summaries rather than across retained states within a chain.

Marker fidelity was the fraction of generated sequences with the designated residue at the
marker position. Attention diagnostics recorded the designated attention mass
\begin{equation}\label{eq:a-des}
  \bar{a}_{\mathrm{des}}
  = \frac{1}{|\mathcal{T}|}\sum_{t \in \mathcal{T}}\;\sum_{k \in \mathcal{D}} a_k(\vxi_t),
  \qquad
  a(\vxi) = \softmax\!\bigl(\beta\,\vX^{\!\top}\vxi + \log\vr\bigr),
\end{equation}
the attention weight summed over the designated index set $\mathcal{D}$
($|\mathcal{D}| = K_{\mathrm{des}}$) and averaged over the retained states $\mathcal{T}$.
Sequence diversity was one minus mean pairwise sequence identity, estimated from 300--500
random pairs drawn with a replicate-local RNG, and amino acid composition fidelity was
$D_{\mathrm{KL}}(\text{reference}\parallel\text{generated})$. The separation
index $S$ was computed from pairwise cosine similarities between unit-norm PCA vectors within
and between designated and background groups. Structural validation used
ESMFold~\cite{esmfold2022} to predict structures for 50 sequences per source. We averaged
pLDDT over populated atom records in the output B-factor column and computed TM-scores against experimental reference
structures (1BPI chain~A for Kunitz and 1OMG chain~A for $\omega$-conotoxin) using
TM-align~\cite{tmAlign2005}. We also ran AlphaFold2-ptm through
ColabFold~\cite{jumperHighlyAccurate2021,mirdita2022colabfold} with one model, three recycles,
an MMseqs2 MSA, and no relaxation, on a Google Colab T4 GPU. Both predictors scored the same
sequence sets, allowing source-group patterns to be compared across predictors
(Table~\ref{tab:structure-model-comparison}). Sequence
plausibility was assessed with ESM2-650M pseudo-perplexity from masked marginal scoring over all
positions~\cite{meierZeroShot2021}, using the ESM2 model~\cite{esmfold2022}. Formal within-chain effective sample sizes and convergence
diagnostics were not used, and instead, at $\rho\in\{1,10,500\}$, we compared
1{,}640 retained ULA states with 1{,}640 independent draws from the exact
Gaussian-mixture target at the same $\beta^{*}$. Exact states were generated by drawing
$k$ from the normalized multiplicities and then drawing
$\vxi\sim\mathcal{N}(\vm_k,\beta^{-1}\mathbf{I})$, and both sets of states were passed
through the same PCA reconstruction and argmax decoder. We defined MAP-designated basin
occupancy as the fraction of states whose maximum-posterior component belonged to the
designated set, and unlike $f_{\mathrm{eff}}$, this hard assignment is not a mixture weight.
This benchmark measures the combined finite-time and discretization error of the ULA
implementation~\cite{durmusMoulinesULA2017}.

Conditioning enters the sampler as an additive bias $\mathbf{b}$ on the attention logits:
\[
\mathrm{softmax}(\beta\,X^{\top}\xi + \mathbf{b}),
\qquad b_k \in \mathbb{R} \cup \{-\infty\}.
\]
The unweighted sampler corresponds to $\mathbf{b} = \mathbf{0}$, multiplicity weighting to
$\mathbf{b} = \log\vr$, and hard attention masking to $b_k = -\infty$ on background patterns,
which assigns those memories softmax weight exactly zero. Because the softmax is invariant to a
constant shift of all logits, multiplicity weighting with ratio $\rho$ is, up to such a shift, a
soft background mask with bias $-\log\rho$, and the hard mask is its $\rho \to \infty$ limit at
fixed $\beta$~\cite{varnerSAProtein2026}. We evaluated the hard mask on the Kunitz family using
the designated K/R-at-P1 split ($K_{\mathrm{des}} = 32$), retaining the full-family PCA basis so
that background patterns receive zero attention while the $d = 80$ coordinate system is unchanged.
This isolates the effect of attention restriction from the basis rebuilding used by hard
curation. Every condition used 30 independent Langevin chains whose initial patterns were
matched across the mask, multiplicity, and unconditional conditions. We report decoded P1~K/R
fraction and sequence novelty, defined as one minus the identity to the nearest sequence in the
input alignment, as the mean and standard error across chains. We held the mask fixed and swept
$\beta$ over eight logarithmically spaced values in $[2,512]$, and we also evaluated the mask at
each multiplicity condition's $\beta^{*}(\vr)$ to obtain the matched-$\beta$ recovery difference.

\subsection{Matched Profile-HMM Benchmark}

As a matched training-free baseline, we fitted HMMER3 profile HMMs to the same cleaned
alignments used by SA. For each finite multiplicity ratio, every designated alignment row was
assigned the relative HMMER sequence weight $\rho$ and every background row weight 1 through
the Stockholm \texttt{\#=GS WT} annotation. We used \texttt{hmmbuild} with
\texttt{--amino --symfrac 0.0 --wgiven --enone}; thus every retained input column was a match
state, the supplied relative weights were used directly, and no additional effective-sequence
number entropy weighting was applied. The $\rho\in\{1,2,5,10,20,50,100,500\}$ grid matched
the canonical SA sweep. A separate profile HMM fitted only to designated rows represented the
hard-curation endpoint. For every family and condition, we generated five independent
replicates of 620 aligned sequences with \texttt{hmmemit -a}, retained columns identified as
match states by the RF annotation, and computed marker fraction, amino acid composition
divergence, novelty, diversity, gap fraction, and valid-residue fraction in the same canonical
alignment coordinates used for SA. For the Kunitz comparison at $\rho=500$, we additionally
scored the first 50 sequences of the first replicate from each method with the same local
ESM2-650M masked-marginal pseudo-perplexity implementation. The benchmark used
HMMER~3.4~\cite{hmmer3}.
Two unconditional Kunitz baselines were built separately from the matched benchmark. The first
was an unconditional profile HMM fitted to the raw Kunitz seed alignment with default
\texttt{hmmbuild} settings, so it used the default match-state occupancy threshold, the default
sequence weighting, and entropy weighting, none of which were applied in the matched benchmark
above. That baseline carries no designation labels; we emitted 150 sequences per replicate with
\texttt{hmmemit} across five replicates. The two profile-HMM conditions reported in
Table~\ref{tab:structure-validation} are therefore different models and are not interchangeable.
The second was a bootstrap resampling control that drew 150 stored sequences per replicate,
uniformly with replacement from the $K=99$ cleaned Kunitz patterns, over the same five
replicates and with the same metrics. Because every draw is a stored sequence, this control
reproduces the input composition by construction and generates nothing novel, which bounds what
composition fidelity alone can establish.

\subsection{Implementation}

All experiments were implemented in Julia~1.12 using NNlib.jl
(softmax), MultivariateStats.jl (PCA), and Distributions.jl; source code and experiment scripts
are available at \url{https://github.com/varnerlab/SA-Binding-Generation-Study}.

\section*{Acknowledgments}
The author thanks the Cornell Center for Advanced Computing for computational resources.

\paragraph{Author contributions.}
J.D.V.\ conceived the study, developed the theory, wrote the code, performed the experiments, and wrote the paper.

\paragraph{Competing interests.}
The author declares no competing interests.

\paragraph{Data availability.}
The five Pfam seed alignments are publicly available from InterPro under the accession numbers
listed in Materials and Methods. The curated SwissProt $\omega$-conotoxin alignment,
designation audit, canonical per-replicate data, generated tables, matched profile-HMM
emissions, ESM2 scoring outputs, source code, and experiment scripts are available at
\url{https://github.com/varnerlab/SA-Binding-Generation-Study}. No proprietary data were used.

\bibliographystyle{unsrtnat}
\bibliography{References_v1}

\clearpage

\begin{table}[p]
  \centering
  \caption{\textbf{Cross-family summary of all protein families tested.}
  For each family: source database, number of sequences ($K$), aligned length ($L$), PCA
  dimensionality ($d$, capturing 95\% variance), designated subset size ($K_{\mathrm{des}}$),
  sequence readout, designated input fraction ($f_{\mathrm{des}}$), separation index
  ($S$), and observed readout fraction under hard curation
  ($f_{\mathrm{obs}}^{\mathrm{hard}}$). For the five Pfam families, $\Delta$ is the
  calibration gap at $\rho=500$. The conotoxin designation is accession based,
  whereas its readout is Tyr13, so its $\Delta$ compares different quantities.
  All six families were tested with both multiplicity weighting and hard curation.}
  \label{tab:cross-family}
  \small
  \resizebox{\textwidth}{!}{%
  \begin{tabular}{lcccccccccc}
    \toprule
    Family & Source & $K$ & $L$ & $d$ & $K_{\mathrm{des}}$ & Readout
           & $f_{\mathrm{des}}$ & $S$
           & $f_{\mathrm{obs}}^{\mathrm{hard}}$ & $\Delta$ \\
    \midrule
    WW & PF00397 & 420 & 31 & 186 & 69 & Spec. loop & 0.16 & 0.11 & $1.000 \pm 0.000$ & $0.660 \pm 0.016$ \\
Forkhead & PF00250 & 246 & 87 & 172 & 122 & H/N at H3 & 0.50 & 0.17 & $1.000 \pm 0.000$ & $0.267 \pm 0.028$ \\
Kunitz & PF00014 & 99 & 53 & 80 & 32 & P1 K/R & 0.32 & 0.20 & $1.000 \pm 0.000$ & $0.409 \pm 0.029$ \\
SH3 & PF00018 & 55 & 48 & 46 & 33 & Trp & 0.60 & 0.34 & $1.000 \pm 0.000$ & $0.009 \pm 0.003$ \\
Homeobox & PF00046 & 136 & 57 & 101 & 102 & Gln at pos. 50 & 0.75 & 0.42 & $1.000 \pm 0.000$ & $0.054 \pm 0.011$ \\
$\omega$-Conotoxin & SwissProt & 74 & 26 & 34 & 23 & Tyr13 & 0.31 & 0.78 & $0.979 \pm 0.006$ & $0.158 \pm 0.035$ \\
\bottomrule

  \end{tabular}
  }
\end{table}

\begin{table}[t]
  \centering
  \caption{\textbf{Kunitz P1~K/R recovery across the conditioning axis.} Designated
  subset is K/R at the P1 position ($K_{\mathrm{des}} = 32$). All conditions use the
  full-family PCA basis except hard curation, which rebuilds the basis from the
  designated subset. The hard mask ($b = -\infty$ on background) is the
  $\rho \to \infty$ endpoint of the multiplicity axis at fixed $\beta$. Novelty is one
  minus the mean nearest-neighbor identity to the training alignment. P1~K/R is the chain-level
  mean $\pm$ standard error over 30 chains; novelty standard errors are below 0.01.}
  \label{tab:mask-recovery}
  \begin{tabular}{@{}lccc@{}}
    \toprule
    Condition & $\beta$ & P1~K/R $\uparrow$ & Novelty \\
    \midrule
    Unconditional ($b = 0$)                      & 3.8 & $0.39 \pm 0.02$ & 0.42 \\
    Multiplicity ($f_{\mathrm{eff}}=0.99$)       & 7.7 & $0.56 \pm 0.02$ & 0.40 \\
    Hard mask ($b = -\infty$), at $\beta^{*}$    & 3.2 & $0.50 \pm 0.02$ & 0.43 \\
    Hard mask ($b = -\infty$), $\beta = 512$     & 512 & $1.00 \pm 0.00$ & 0.16 \\
    Hard curation (subset basis)                 & 3.2 & $1.00 \pm 0.00$ & 0.32 \\
    \bottomrule
  \end{tabular}
\end{table}

\begin{table}[p]
  \centering
  \caption{\textbf{Per-family $\rho$ sweep: mean attention weights track
  $f_{\mathrm{eff}}$ closely; decoded marker fidelity varies across PCA
  geometries.} Selected $\rho$ values for each family.
  $f_{\mathrm{eff}}$: effective designated fraction from multiplicity weights;
  $\bar{a}_{\mathrm{des}}$: mean softmax attention on designated patterns;
  $f_{\mathrm{obs}}$: mean hard-decoded marker fraction across five independent
  replicates (620 retained samples per replicate); error bars are replicate s.d.;
  $D$: mean pairwise diversity.
  In all families $\bar{a}_{\mathrm{des}} \approx f_{\mathrm{eff}}$
  within finite-sample uncertainty, but $f_{\mathrm{obs}}$ converges fastest
  for Homeobox ($S=\HomeoboxSeparation$) and slowest for WW
  ($S=\WWSeparation$), reflecting the
  observed latent-to-sequence calibration gap. In conotoxin, designation is
  accession based while $f_{\mathrm{obs}}$ measures Tyr13, so the difference is
  descriptive rather than a calibration gap.}
  \label{tab:per-family-rho}
  \small
  \resizebox{\textwidth}{!}{%
  \begin{tabular}{llcccccc}
    \toprule
    Family & ($K$, $K_{\mathrm{des}}$, $S$)
      & $\rho$ & $f_{\mathrm{eff}}$ & $\bar{a}_{\mathrm{des}}$
      & $f_{\mathrm{obs}}$ & $\Delta$ & $D$ \\
    \midrule
    \multirow{4}{*}{WW} & \multirow{4}{*}{(420, 69, 0.11)} & 1 & 0.164 & $0.170 \pm 0.006$ & $0.144 \pm 0.016$ & 0.02 & 0.686 \\
& & 10 & 0.663 & $0.659 \pm 0.013$ & $0.223 \pm 0.028$ & 0.44 & 0.670 \\
& & 100 & 0.952 & $0.952 \pm 0.003$ & $0.291 \pm 0.019$ & 0.66 & 0.652 \\
& & 500 & 0.990 & $0.990 \pm 0.001$ & $0.330 \pm 0.016$ & 0.66 & 0.631 \\
\midrule
\multirow{4}{*}{Forkhead} & \multirow{4}{*}{(246, 122, 0.17)} & 1 & 0.496 & $0.496 \pm 0.006$ & $0.565 \pm 0.013$ & -0.07 & 0.485 \\
& & 10 & 0.908 & $0.907 \pm 0.005$ & $0.612 \pm 0.019$ & 0.30 & 0.468 \\
& & 100 & 0.990 & $0.991 \pm 0.001$ & $0.696 \pm 0.031$ & 0.29 & 0.415 \\
& & 500 & 0.998 & $0.998 \pm 0.001$ & $0.731 \pm 0.028$ & 0.27 & 0.402 \\
\midrule
\multirow{4}{*}{Kunitz} & \multirow{4}{*}{(99, 32, 0.20)} & 1 & 0.323 & $0.327 \pm 0.013$ & $0.382 \pm 0.031$ & -0.06 & 0.557 \\
& & 10 & 0.827 & $0.830 \pm 0.005$ & $0.515 \pm 0.027$ & 0.31 & 0.538 \\
& & 100 & 0.979 & $0.977 \pm 0.002$ & $0.563 \pm 0.023$ & 0.42 & 0.519 \\
& & 500 & 0.996 & $0.996 \pm 0.001$ & $0.587 \pm 0.029$ & 0.41 & 0.510 \\
\midrule
\multirow{4}{*}{SH3} & \multirow{4}{*}{(55, 33, 0.34)} & 1 & 0.600 & $0.602 \pm 0.009$ & $0.839 \pm 0.021$ & -0.24 & 0.586 \\
& & 10 & 0.938 & $0.935 \pm 0.003$ & $0.900 \pm 0.012$ & 0.04 & 0.581 \\
& & 100 & 0.993 & $0.995 \pm 0.001$ & $0.985 \pm 0.006$ & 0.01 & 0.524 \\
& & 500 & 0.999 & $0.999 \pm 0.001$ & $0.989 \pm 0.003$ & 0.01 & 0.515 \\
\midrule
\multirow{4}{*}{Homeobox} & \multirow{4}{*}{(136, 102, 0.42)} & 1 & 0.750 & $0.747 \pm 0.016$ & $0.931 \pm 0.017$ & -0.18 & 0.523 \\
& & 10 & 0.968 & $0.967 \pm 0.001$ & $0.948 \pm 0.004$ & 0.02 & 0.520 \\
& & 100 & 0.997 & $0.996 \pm 0.000$ & $0.927 \pm 0.015$ & 0.07 & 0.535 \\
& & 500 & 0.999 & $0.999 \pm 0.000$ & $0.945 \pm 0.011$ & 0.05 & 0.536 \\
\midrule
\multirow{4}{*}{$\omega$-Conotoxin} & \multirow{4}{*}{(74, 23, 0.78)} & 1 & 0.311 & $0.304 \pm 0.025$ & $0.463 \pm 0.040$ & -0.15 & 0.560 \\
& & 10 & 0.819 & $0.819 \pm 0.013$ & $0.662 \pm 0.029$ & 0.16 & 0.510 \\
& & 100 & 0.978 & $0.982 \pm 0.002$ & $0.800 \pm 0.020$ & 0.18 & 0.460 \\
& & 500 & 0.996 & $0.996 \pm 0.002$ & $0.838 \pm 0.035$ & 0.16 & 0.443 \\
\bottomrule

  \end{tabular}
  }
\end{table}

\begin{table}[p]
  \centering
  \caption{\textbf{Matched multiplicity-weighted profile-HMM benchmark at
  $\rho=500$.}
  Both methods receive the same cleaned alignment, designation labels, and relative
  sequence weights. Marker fractions are mean $\pm$ s.d.\ across five independent
  replicates of 620 sequences. $D$ is mean pairwise sequence diversity. The profile
  HMM transfers the selected single-position marginal directly in the five Pfam splits;
  SA additionally passes through its PCA decoder. For conotoxin, $f_{\mathrm{eff}}$
  refers to accession designation and the marker column reports Tyr13.}
  \label{tab:profile-hmm-benchmark}
  \small
  \begin{tabular}{lccccc}
    \toprule
    Family & $f_{\mathrm{eff}}$ & SA marker & HMM marker & SA $D$ & HMM $D$ \\
    \midrule
    Kunitz & 0.996 & $0.587 \pm 0.029$ & $0.996 \pm 0.002$ & 0.510 & 0.590 \\
SH3 & 0.999 & $0.989 \pm 0.003$ & $0.999 \pm 0.001$ & 0.515 & 0.655 \\
WW & 0.990 & $0.330 \pm 0.016$ & $0.988 \pm 0.004$ & 0.631 & 0.628 \\
Homeobox & 0.999 & $0.945 \pm 0.011$ & $1.000 \pm 0.001$ & 0.536 & 0.605 \\
Forkhead & 0.998 & $0.731 \pm 0.028$ & $0.997 \pm 0.001$ & 0.402 & 0.562 \\
$\omega$-Conotoxin & 0.996 & $0.838 \pm 0.035$ & $0.834 \pm 0.013$ & 0.443 & 0.381 \\
\bottomrule

  \end{tabular}
\end{table}

\begin{table}[p]
  \centering
  \caption{\textbf{Structure-model, language-model, and sequence-marker evaluation on the Kunitz domain.}
  pLDDT: ESMFold predicted confidence (0--100; $>70$ = confident).
  TM: TM-score against 1BPI chain~A ($>0.5$ is a conventional same-fold reference).
  PPL: ESM2-650M masked-marginal pseudo-perplexity (lower = more plausible).
  P1~K/R: fraction with Lys/Arg at the primary trypsin contact.
  KL: amino acid composition divergence ($\times 10^{3}$).
  $D$: mean pairwise diversity.
  Structural and LM values are mean $\pm$ s.d.\ across 50 sequences;
  sequence-level metrics are mean $\pm$ s.d.\ across 5 replicates.
  Structure-model scores were not computed for the additional weighted profile-HMM
  benchmark.
  Sequences generated by SA have structure-model and language-model summaries similar to
  those of stored sequences. Conditioning on the K/R-positive subset yields complete K/R marker retention;
  this is not an experimental binding measurement.}
  \label{tab:structure-validation}
  \small
  \resizebox{\textwidth}{!}{%
  \begin{tabular}{lcccccc}
    \toprule
    Source & pLDDT & TM & PPL & P1 K/R & KL ($\times 10^{3}$) & $D$ \\
    \midrule
    Stored (natural)      & $89.3 \pm 3.2$  & $0.83 \pm 0.02$ & $4.97 \pm 0.98$ & 0.32 & 0 & 0.63 \\
    SA (full family)      & $90.4 \pm 1.2$  & $0.84 \pm 0.01$ & $4.78 \pm 0.64$ & $0.38 \pm 0.03$ & $7.6 \pm 0.9$ & 0.56 \\
    SA (K/R-positive)     & $90.7 \pm 1.1$  & $0.83 \pm 0.01$ & $4.72 \pm 0.60$ & $1.00 \pm 0.00$ & $18.7 \pm 0.9$ & 0.50 \\
    SA (K/R-negative)     & $91.0 \pm 1.2$  & $0.85 \pm 0.01$ & $4.73 \pm 0.72$ & $0.00 \pm 0.00$ & $11.2 \pm 0.2$ & 0.53 \\
    HMM emit (HMMER3)     & $60.4 \pm 13.1$ & $0.60 \pm 0.19$ & $16.1 \pm 4.2$  & $0.17 \pm 0.04$ & $27.1 \pm 2.1$ & 0.90 \\
    HMM weighted ($\rho=500$) & --- & --- & $\HMMRhoFiveHundredPPL \pm \HMMRhoFiveHundredPPLSD$ & $\HMMRhoFiveHundredMarker \pm \HMMRhoFiveHundredMarkerSD$ & $\HMMRhoFiveHundredKL \pm \HMMRhoFiveHundredKLSD$ & \HMMRhoFiveHundredDiversity{} \\
    Bootstrap (resample)  & ---             & ---              & ---              & $0.30 \pm 0.04$ & $1.0 \pm 0.3$ & 0.62 \\
    \bottomrule
  \end{tabular}
  }
\end{table}

\begin{table}[p]
  \centering
  \caption{\textbf{Cross-predictor structure-model comparison for Kunitz and
  $\omega$-conotoxin.}
  Two structure predictors scored the same 50 sequences per source:
  AlphaFold2-ptm via ColabFold (1 model, 3 recycles, MMseqs2 MSA) and ESMFold.
  pLDDT: mean predicted confidence ($0$--$100$; $>70$ = confident).
  TM: TM-score against the experimental reference (1BPI chain~A for Kunitz,
  1OMG chain~A for $\omega$-conotoxin; $>0.5$ is a conventional same-fold reference).
  Values are mean $\pm$ s.d.\ across the 50 sequences.
  The designated-subset $\omega$-conotoxins had a higher mean TM-score than the stored group
  under both predictors. For full-family generation, the ESMFold mean was higher, whereas the
  AlphaFold2-ptm mean exceeded the stored mean by only 0.0027, negligible relative to the
  within-group standard deviations. For $\omega$-conotoxin, the overlap between stored
  and generated groups gives no evidence of a conditioning-induced loss. The low absolute
  TM-scores alone cannot distinguish short-chain metric behavior from structural
  differences. These model-derived quantities do not establish experimental folding or
  binding.}
  \label{tab:structure-model-comparison}
  \small
  \begin{tabular}{llccccc}
    \toprule
    & & & \multicolumn{2}{c}{AlphaFold2-ptm} & \multicolumn{2}{c}{ESMFold} \\
    \cmidrule(lr){4-5}\cmidrule(lr){6-7}
    Family & Source & $n$ & pLDDT & TM & pLDDT & TM \\
    \midrule
    \multirow{3}{*}{Kunitz}
      & Stored (natural)       & 50 & $93.5 \pm 4.0$  & $0.83 \pm 0.03$ & $89.3 \pm 3.2$ & $0.83 \pm 0.02$ \\
      & SA (K/R-positive)      & 50 & $94.6 \pm 1.4$  & $0.84 \pm 0.01$ & $90.7 \pm 1.1$ & $0.83 \pm 0.01$ \\
      & SA (full family)       & 50 & $95.0 \pm 1.3$  & $0.85 \pm 0.01$ & $90.4 \pm 1.2$ & $0.84 \pm 0.01$ \\
    \midrule
    \multirow{3}{*}{$\omega$-Conotoxin}
      & Stored (natural)       & 50 & $67.9 \pm 9.8$  & $0.35 \pm 0.11$ & $76.8 \pm 7.7$ & $0.43 \pm 0.07$ \\
      & SA (designated subset) & 50 & $79.5 \pm 4.0$  & $0.47 \pm 0.02$ & $78.1 \pm 2.5$ & $0.47 \pm 0.03$ \\
      & SA (full family)       & 50 & $70.0 \pm 11.1$ & $0.35 \pm 0.13$ & $78.1 \pm 3.3$ & $0.47 \pm 0.03$ \\
    \bottomrule
  \end{tabular}
\end{table}

\begin{table}[p]
  \centering
  \caption{\textbf{$\omega$-Conotoxin Tyr13 marker retention under curated vs.\
  full-family seeding.}
  Tyr13\%: fraction with tyrosine at the selected marker position (col.~13).
  K+R\%: overall basic-residue fraction. KL: amino acid composition
  divergence. Novelty: mean cosine distance to nearest stored pattern. Designated-subset seeding
  increased the Tyr13 frequency from 82.6\% to 98.5\% and produced a similar
  basic-residue fraction (19.7\% against 20.1\%). Full-family seeding produced lower values than
  designated-subset seeding, although both Tyr13 and the basic-residue fraction increased relative
  to the full-family input.}
  \label{tab:conotoxin-pharmacophore}
  \small
  \begin{tabular}{lccccc}
    \toprule
    Condition & $N$ & Tyr13 (\%) & K+R (\%) & KL & Novelty \\
    \midrule
    Input: full family          &   74 & 33.8 &  9.9 & 0      & 0 \\
    Input: designated subset    &   23 & 82.6 & 19.7 & 0      & 0 \\
    Generated: full-seeded      & 1550 & 48.5 & 12.0 & 0.0084 & 0.59 \\
    Generated: designated-seeded & 1550 & 98.5 & 20.1 & 0.0084 & 0.45 \\
    \bottomrule
  \end{tabular}
\end{table}

\begin{table}[ht]
  \caption{\textbf{Residue frequencies at reported $\omega$-conotoxin SAR positions.}
  Published SAR and structural studies across $\omega$-conotoxins report roles for
  these positions in activity, binding, or fold
  maintenance~\cite{kimTyr13Essential1995,satoBasicResidues1993,satoAlaScanMVIIC2000,schroederLys2Effects2012,lewisConotoxinSAR2012,schroederChiralityTyr131999}.
  For each reported position, the table gives the fraction retaining the listed residue
  (or a K/R substitution at basic positions). The six framework cysteines are invariant
  across the designated input, so the table reports their retention rather than their
  enrichment. Designated-subset generation retains the Tyr13 signal and the cysteine
  scaffold, and increases the listed frequencies at Tyr13, Arg10, position~21, and
  Lys4 relative to the designated input. These
  frequencies do not test the effects of mutations in the generated backgrounds.}
  \label{tab:sar-agreement}
  \centering
  \small
  \begin{tabular}{clllccc}
    \toprule
    Pos. & WT & Reported role & Evidence & Input & SA des. & SA full \\
    \midrule
    13 & Tyr & Reported Tyr marker & Ala: large activity loss & 0.83 & \textbf{0.98} & 0.48 \\
2 & Lys & Loop 2 stabilization & Ala: 40$\times$ loss (GVIA) & 0.96 & \textbf{0.98} & 0.45 \\
10 & Arg & Reported loop 2 residue & MVIIA binding region & 0.70 & \textbf{0.83} & 0.36 \\
11 & Leu & Reported loop 2 residue & MVIIA binding region & 0.17 & 0.20 & 0.13 \\
1 & Cys & Cysteine framework & Disulfide-forming & 1.00 & \textbf{1.00} & 1.00 \\
8 & Cys & Cysteine framework & Disulfide-forming & 1.00 & \textbf{1.00} & 1.00 \\
15 & Cys & Cysteine framework & Disulfide-forming & 1.00 & \textbf{1.00} & 1.00 \\
16 & Cys & Cysteine framework & Disulfide-forming & 1.00 & \textbf{1.00} & 1.00 \\
20 & Cys & Cysteine framework & Disulfide-forming & 1.00 & \textbf{1.00} & 1.00 \\
25 & Cys & Cysteine framework & Disulfide-forming & 1.00 & \textbf{1.00} & 1.00 \\
21 & Arg & Reported MVIIA SAR & Ala: reduced potency & 0.48 & \textbf{0.52} & 0.14 \\
4 & Lys & MVIIC P/Q selectivity & Ala: affects P/Q & 0.52 & \textbf{0.60} & 0.27 \\
\bottomrule

  \end{tabular}
\end{table}

\clearpage

\begin{figure}[p]
  \centering
  \includegraphics[width=0.82\textwidth]{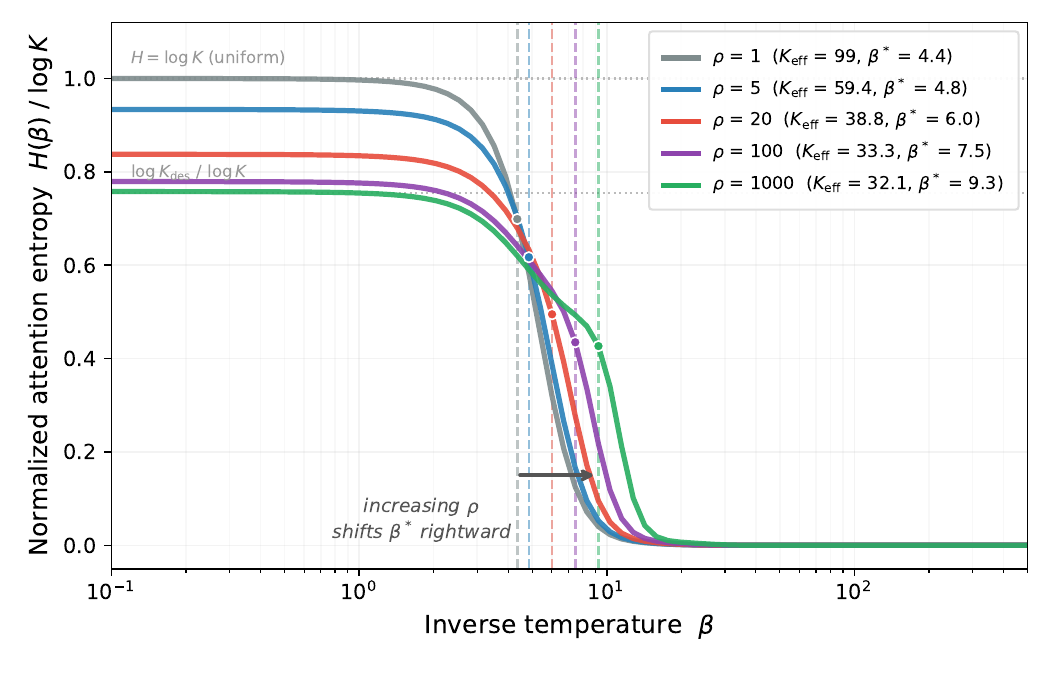}
  \caption{\textbf{Entropy crossover shifts rightward with increasing multiplicity
  ratio $\rho$.}
  Attention entropy $H(\beta)$ normalized by $\log K$ (the uniform limit) as a
  function of inverse temperature $\beta$ (log scale) for
  $\rho \in \{1, 5, 20, 100, 1000\}$ on the Kunitz family ($K=99$,
  $K_{\mathrm{des}}=32$). At $\rho=1$ (standard SA), the entropy starts at
  $H/\log K = 1$ (uniform attention over all patterns); as $\rho$ increases,
  the multiplicity bias pre-concentrates the attention distribution, lowering the
  starting entropy toward $\log K_{\mathrm{des}} / \log K \approx 0.75$ (dotted
  line). Each curve shows a sigmoidal drop to near-zero entropy; dots and dashed
  vertical lines mark the crossover onset $\beta^{*}(\rho)$, the point of maximum
  downward curvature. As $\rho$ increases from 1 to 1000, $K_{\mathrm{eff}}$
  decreases from 99 to 32.1 and $\beta^{*}$ shifts rightward from 4.4 to 9.3.}
  \label{fig:phase-transition}
\end{figure}

\clearpage
\setcounter{figure}{0}
\setcounter{table}{0}
\renewcommand{\thefigure}{S\arabic{figure}}
\renewcommand{\thetable}{S\arabic{table}}
\setcounter{section}{0}
\renewcommand{\thesection}{S\arabic{section}}
\renewcommand{\theHfigure}{S\arabic{figure}}
\renewcommand{\theHtable}{S\arabic{table}}
\renewcommand{\theHsection}{S\arabic{section}}

\section*{SI Appendix}
\providecommand{\PaperRoot}{.}
\providecommand{\MainRef}[1]{\ref{#1}}
\providecommand{\MainEqRef}[1]{\eqref{#1}}

\section{Sequence Encoding and Multiplicity Conditioning}\label{app:pca}

Stochastic attention (SA) stores unit-norm memory vectors and evolves continuous
states in $\mathbb{R}^d$, while protein sequences are discrete objects over a
20-letter amino acid alphabet. Bridging the two requires three transformations
(one-hot encoding, PCA dimensionality reduction, and unit-norm projection), and the reverse path
(affine PCA reconstruction followed by argmax decoding) maps continuous samples
to discrete sequences. Here we describe the full encoding--decoding pipeline,
derive the multiplicity-weighted conditioning formulas used in the main text,
and discuss how the multiplicity vector $\mathbf{r}$ interacts with the PCA
representation.

Given a cleaned alignment of $K$ sequences, each of length $L$ amino acids
over the standard 20-letter alphabet
$\mathcal{A} = \{\text{A}, \text{R}, \text{N}, \ldots, \text{V}\}$, we
encoded each sequence as a binary vector in $\mathbb{R}^{20L}$ by
concatenating per-position indicator vectors. For sequence $k$ ($k = 1,
\ldots, K$) with amino acid $a_\ell^{(k)}$ at alignment position $\ell$
($\ell = 1, \ldots, L$):
\begin{equation}
  \mathbf{x}_k = \bigl[\mathbf{e}_{a_1^{(k)}},\; \mathbf{e}_{a_2^{(k)}},\;
    \ldots,\; \mathbf{e}_{a_L^{(k)}}\bigr]^\top \in \{0,1\}^{20L},
\end{equation}
where $\mathbf{e}_a \in \{0,1\}^{20}$ is the standard basis vector for amino
acid $a$ (i.e., a one-hot indicator with a 1 in the position corresponding to
$a$ and 0 elsewhere). The full one-hot dimensionality is
$d_{\mathrm{full}} = 20L$, which ranged from 520 ($\omega$-conotoxin,
$L{=}26$) to 1{,}740 (Forkhead, $L{=}87$) across the six families. This space
is over-dimensioned relative to the number of stored patterns $K$
($d_{\mathrm{full}}/K$ ranged from 1.48 to 17.45), which creates a problem for
the modern Hopfield energy~\cite{ramsauerHopfieldNetworksAll2021}: the
similarity score $e_k = \mathbf{m}_k^\top\boldsymbol{\xi}$ between a stored
pattern $\mathbf{m}_k$ and a query state $\boldsymbol{\xi}$ has variance
$1/d$ for a random query on the unit sphere $\mathbb{S}^{d-1}$, so when $d$
is large all scores concentrate near zero and the inverse temperature $\beta$
must be set extremely high to differentiate among patterns, making Langevin
sampling inefficient. Principal component analysis removes this redundancy
by projecting onto the $d \ll d_{\mathrm{full}}$ directions of actual
variation in the alignment.

We computed PCA from the one-hot encoded alignment by first centering the data:
$\bar{\mathbf{x}} = K^{-1}\sum_{k=1}^K \mathbf{x}_k$ (the family's
position-specific amino acid frequency profile),
$\tilde{\mathbf{X}} = \mathbf{X} - \bar{\mathbf{x}}\mathbf{1}_K^\top$,
and then computing the economy singular value decomposition (SVD)
$\tilde{\mathbf{X}} = \mathbf{U}\boldsymbol{\Sigma}\mathbf{V}^\top$, where
$\mathbf{U} \in \mathbb{R}^{d_{\mathrm{full}} \times r}$ contains the
principal directions of variation,
$\boldsymbol{\Sigma} = \mathrm{diag}(\sigma_1, \ldots, \sigma_r)$ the
singular values in decreasing order, and
$r = \mathrm{rank}(\tilde{\mathbf{X}}) \leq \min(d_{\mathrm{full}}, K) - 1$.
We retained the smallest dimension $d$ such that
$\sum_{j=1}^d \sigma_j^2 / \sum_{j=1}^r \sigma_j^2 \geq 0.95$ (at least
95\% of total variance). Across the six families, $d$ ranged from 34
($\omega$-conotoxin) to 186 (WW), yielding compression ratios of
$3.3\times$ to $21\times$ (Table~\MainRef{tab:cross-family}). Each sequence was
then projected as
$\mathbf{z}_k = \mathbf{W}_d^\top(\mathbf{x}_k - \bar{\mathbf{x}})
\in \mathbb{R}^d$, where
$\mathbf{W}_d = \mathbf{U}_{:,1:d}$ is the projection matrix, and
normalized to unit $L_2$ norm:
$\mathbf{m}_k = \mathbf{z}_k / \|\mathbf{z}_k\|_2 \in \mathbb{S}^{d-1}$.
This normalization ensures that similarity scores lie in $[-1, 1]$ and
prevents patterns with larger PCA norms from dominating the softmax attention.
The columns $\mathbf{m}_1, \ldots, \mathbf{m}_K$ form the memory matrix
$\hat{\mathbf{X}} \in \mathbb{R}^{d \times K}$ used throughout the paper.
The Langevin sampler (Eq.~\MainRef{eq:ula}) produces continuous states
$\boldsymbol{\xi} \in \mathbb{R}^d$, which are decoded back to amino acid
sequences by affine PCA reconstruction
$\hat{\mathbf{x}} = \bar{\mathbf{x}} + \mathbf{W}_d\boldsymbol{\xi}
\in \mathbb{R}^{d_{\mathrm{full}}}$ followed by per-position argmax:
$a_\ell = \arg\max_{a \in \mathcal{A}}\; \hat{x}_{20(\ell-1) + a}$ for
$\ell = 1, \ldots, L$. This decoding is deterministic and produced valid
amino acids at 100\% of positions across all families and conditions. It is
not an inverse of the complete encoder: unit normalization discards each
stored score radius $\|\mathbf{z}_k\|_2$, and the decoder does not restore it.
For stored memories, reconstructing from
$\|\mathbf{z}_k\|_2\mathbf{m}_k$ increases mean sequence identity from
0.730--0.943 to 0.996--1.000 across families. A generated sample, however, is
not a normalized encoding of a source sequence and has no known discarded
radius. Multiplying it by a stored-memory radius scales both its component
signal and its finite-temperature noise, so we treat such rescaling as a
decoder-sensitivity intervention rather than an inverse operation.

The multiplicity-weighted Hopfield energy (Eq.~\MainRef{eq:weighted-energy})
assigns a weight $r_k > 0$ to each stored pattern $\mathbf{m}_k$. For the
binary functional split used throughout this paper, we set $r_k = \rho$ for
the $K_{\mathrm{des}}$ designated patterns and $r_k = 1$ for the
$K_{\mathrm{bg}} = K - K_{\mathrm{des}}$ background patterns, where
$\rho \geq 1$ is the multiplicity ratio. The weighted score function
(Proposition~\MainRef{prop:score}) adds $\log\rho$ to the pre-softmax logits of
designated patterns and $\log 1 = 0$ to background patterns. To derive the
effect of $\rho$ on the attention distribution, consider the idealized case
where the query $\boldsymbol{\xi}$ has equal similarity to all stored
patterns, i.e., $\mathbf{m}_k^\top\boldsymbol{\xi} = c$ for all $k$. The
softmax attention weight on pattern $k$ simplifies to
$a_k = r_k / \sum_{j} r_j$, and the total attention weight on all designated
patterns is given by:
\begin{equation}\label{eq:feff-def}
  f_{\mathrm{eff}}(\rho)
  \;=\; \sum_{k \in \mathrm{des}} a_k
  \;=\; \frac{K_{\mathrm{des}}\,\rho}{K_{\mathrm{des}}\,\rho + K_{\mathrm{bg}}}.
\end{equation}
At $\rho = 1$, $f_{\mathrm{eff}} = K_{\mathrm{des}} / K$ (the natural
proportion); as $\rho \to \infty$, $f_{\mathrm{eff}} \to 1$. To find the
multiplicity ratio needed for a target effective fraction $f \in (0, 1)$, we
invert Eq.~\eqref{eq:feff-def} by multiplying both sides by the denominator
and solving for $\rho$:
\begin{align}
  f\,(K_{\mathrm{des}}\,\rho + K_{\mathrm{bg}}) &= K_{\mathrm{des}}\,\rho,
  \notag\\
  f\,K_{\mathrm{bg}} &= K_{\mathrm{des}}\,\rho\,(1 - f), \notag\\
  \rho(f) &= \frac{f\,K_{\mathrm{bg}}}{K_{\mathrm{des}}\,(1-f)}.
  \label{eq:rho-inverse-derived}
\end{align}
This is Eq.~\MainEqRef{eq:rho-inverse} in the main text. Equal similarity gives
Eq.~\eqref{eq:feff-def} pointwise. More generally, the exact Gaussian-mixture
identity (Appendix~\ref{app:gmm}) makes $f_{\mathrm{eff}}$ the exact latent
designated-component probability, and the expected posterior designated
attention equals the same quantity by the law of total expectation. In the
finite-step canonical ULA sweeps, the largest absolute deviation of empirical
mean attention from $f_{\mathrm{eff}}(\rho)$ was \AttnMaxDevPts{} percentage
points (\AttnMaxDevFamily{}, $\rho=\AttnMaxDevRho$;
Table~\MainRef{tab:per-family-rho}). This residual includes finite sampling, mixing,
and ULA discretization error. The effective number of patterns under
multiplicity weighting,
$K_{\mathrm{eff}}(\mathbf{r}) = (\sum_k r_k)^2 / \sum_k r_k^2
= (K_{\mathrm{des}}\,\rho + K_{\mathrm{bg}})^2 /
(K_{\mathrm{des}}\,\rho^2 + K_{\mathrm{bg}})$, equals $K$ when $\rho = 1$
and converges to $K_{\mathrm{des}}$ as $\rho \to \infty$. It is a descriptive
summary of the weight concentration and is reported alongside the observed shift in
$\beta^{*}$, not as its cause.

A key property of multiplicity-weighted conditioning is that $\mathbf{r}$
does \emph{not} alter the PCA encoding. The memory matrix
$\hat{\mathbf{X}}$ is constructed once from the full family alignment and
remains fixed regardless of $\rho$; the multiplicity weights enter only
through the softmax logits. This separation has three consequences.
First, the PCA subspace is independent of $\rho$: the principal directions
$\mathbf{W}_d$, the centering vector $\bar{\mathbf{x}}$, and the retained
dimension $d$ are all determined by the alignment alone. This fixes the
representation through which the conditioning signal must pass, but it does
not fix the decoded marker response: both the latent share and the response
vary with $\rho$ in the replicated sweeps. For the five Pfam families,
designation and marker status coincide, so their difference is a calibration
gap. The conotoxin designation is accession based and does not coincide with
Tyr13 status; its analogous difference is descriptive. Second,
$\rho$ controls attention, not
geometry: increasing $\rho$ adds $\log\rho$ to the designated logits, shifting
attention weight toward designated patterns, but this shift must still pass
through finite-temperature latent noise, the normalized-coordinate PCA
reconstruction, and argmax decoding to produce discrete residues. These
operations jointly determine the decoded response. Third, $\rho$ shifts the
entropy crossover: the logit biases make the attention uneven at $\beta = 0$
before $\beta$ does anything, reducing the weighted entropy there to the Shannon
entropy $H_{\mathbf{r}}(0) = -\sum_k w_k \log w_k < \log K$ with
$w_k = r_k / \sum_j r_j$. We observed that the crossover moved to higher inverse
temperature as $\rho$ increased:
$\beta^{*}$ increased from 4.4 ($\rho=1$, $K_{\mathrm{eff}} = 99$) to 9.3
($\rho=1{,}000$, $K_{\mathrm{eff}} = 32.1$) on the Kunitz domain
(Fig.~\MainRef{fig:phase-transition}). By contrast, hard curation (where the
memory matrix is built from the designated subset only) \emph{does} change
the PCA: the centering vector, principal directions, and retained dimension
are all computed from $K_{\mathrm{des}}$ sequences, capturing
subset-specific variation that may be orthogonal to the leading components
of the full-family PCA. Hard curation achieved 100\% marker transfer in the
five Pfam families and 97.9\% in conotoxin, whereas multiplicity weighting
retained the full-family PCA geometry.

\section{Supplementary Tables and Figures}\label{app:tables-figures}

\begin{figure}[p]
  \centering
  \includegraphics[width=0.72\textwidth]{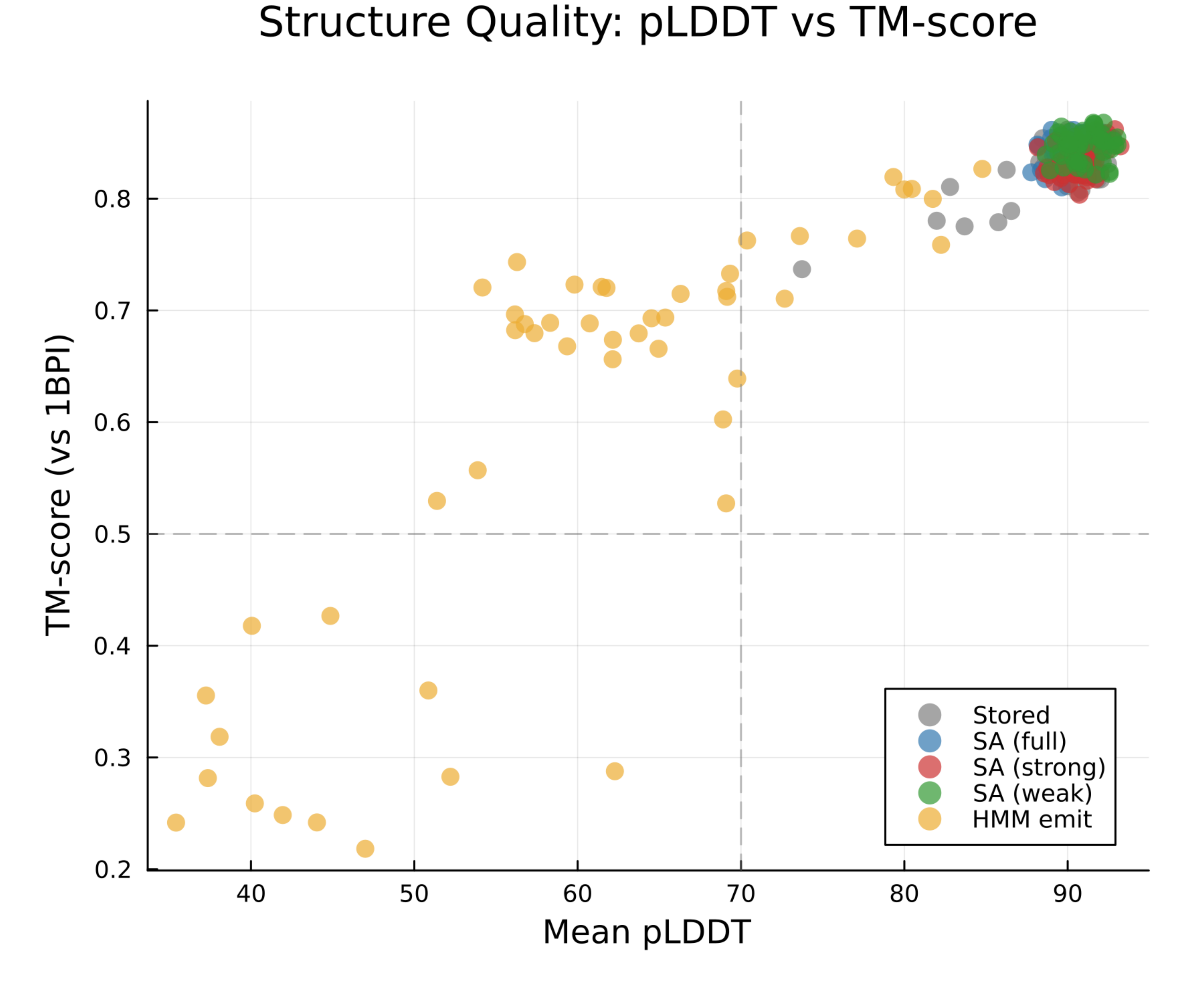}
  \caption{\textbf{Joint pLDDT and TM-score distributions for Kunitz sequences.}
  Each point represents one predicted structure (50 per source). SA-generated
  (blue, red, green) and most stored (gray) sequences occupy the high-score region near
  pLDDT $>85$ and TM-score $>0.8$. Profile HMM sequences (orange) scatter broadly, with many falling below the
  pLDDT $>70$ confidence reference (vertical dashed line) and the conventional TM-score
  $>0.5$ same-fold reference (horizontal dashed line).}
  \label{fig:plddt-tmscore-scatter}
\end{figure}

Tables~\ref{tab:kunitz-rho}--\ref{tab:conotoxin-rho} report the canonical
replicated $\rho$ sweep on the five Pfam benchmark families and
$\omega$-conotoxin.
$f_{\mathrm{eff}}$ is the effective designated fraction computed from the
multiplicity vector; $f_{\mathrm{obs}}$ is the hard-decoded marker fraction
averaged across five independent replicates (620 retained samples per replicate);
$\bar{a}_{\mathrm{des}}$ is the softmax attention weight summed over the designated
patterns and averaged over retained states; and $D$ is the mean pairwise
diversity. Uncertainties are replicate
standard deviations. For the five Pfam families, designated sequences are
exactly the marker-positive sequences. For conotoxin, designation follows
curated accession membership, so $f_{\mathrm{eff}}-f_{\mathrm{obs}}$ compares
an accession share with a Tyr13 readout and is not a calibration error between
like quantities. Marker fraction and diversity are reported as a function
of $\beta/\beta^{*}$ for three fixed $\rho$ values on the Kunitz domain
(Table~\ref{tab:beta-sweep}). Across all 250 predicted Kunitz structures,
SA-generated sequences cluster tightly in the high-pLDDT, high-TM-score region
of the joint ESMFold pLDDT and TM-score distribution while HMM-emitted
sequences scatter broadly across lower values
(Fig.~\ref{fig:plddt-tmscore-scatter}). The entropy curves for the full $\rho$ sweep
($\rho \in \{1, 2, 5, 10, 20, 50, 100, 200, 500, 1000\}$) on the Kunitz
family produce a monotonically rightward-shifting family of sigmoid curves
with the operating point increasing from 4.35 ($\rho=1$) to 9.26 ($\rho=1000$)
(Fig.~\MainRef{fig:phase-transition}). We report this displacement without asserting a
functional relationship to $K_{\mathrm{eff}}$.

\begin{table}[ht]
  \centering
  \caption{Kunitz domain (PF00014) $\rho$ sweep. $K=99$, $K_{\mathrm{des}}=32$.}
  \label{tab:kunitz-rho}
  \small
  \begin{tabular}{ccccc}
    \toprule
    $\rho$ & $f_{\mathrm{eff}}$ & $f_{\mathrm{obs}}$ & $\bar{a}_{\mathrm{des}}$ & $D$ \\
    \midrule
    1 & 0.323 & $0.382 \pm 0.031$ & $0.327 \pm 0.013$ & $0.557 \pm 0.004$ \\
2 & 0.489 & $0.412 \pm 0.031$ & $0.477 \pm 0.009$ & $0.556 \pm 0.003$ \\
5 & 0.705 & $0.481 \pm 0.014$ & $0.712 \pm 0.005$ & $0.548 \pm 0.004$ \\
10 & 0.827 & $0.515 \pm 0.027$ & $0.830 \pm 0.005$ & $0.538 \pm 0.003$ \\
20 & 0.905 & $0.504 \pm 0.015$ & $0.904 \pm 0.006$ & $0.540 \pm 0.001$ \\
50 & 0.960 & $0.554 \pm 0.031$ & $0.960 \pm 0.005$ & $0.531 \pm 0.003$ \\
100 & 0.979 & $0.563 \pm 0.023$ & $0.977 \pm 0.002$ & $0.519 \pm 0.003$ \\
500 & 0.996 & $0.587 \pm 0.029$ & $0.996 \pm 0.001$ & $0.510 \pm 0.002$ \\
\bottomrule

  \end{tabular}
\end{table}

\begin{table}[ht]
  \centering
  \caption{SH3 domain (PF00018) $\rho$ sweep. $K=55$, $K_{\mathrm{des}}=33$.}
  \label{tab:sh3-rho}
  \small
  \begin{tabular}{ccccc}
    \toprule
    $\rho$ & $f_{\mathrm{eff}}$ & $f_{\mathrm{obs}}$ & $\bar{a}_{\mathrm{des}}$ & $D$ \\
    \midrule
    1 & 0.600 & $0.839 \pm 0.021$ & $0.602 \pm 0.009$ & $0.586 \pm 0.004$ \\
2 & 0.750 & $0.861 \pm 0.011$ & $0.747 \pm 0.005$ & $0.582 \pm 0.005$ \\
5 & 0.882 & $0.899 \pm 0.013$ & $0.889 \pm 0.011$ & $0.579 \pm 0.002$ \\
10 & 0.938 & $0.900 \pm 0.012$ & $0.935 \pm 0.003$ & $0.581 \pm 0.003$ \\
20 & 0.968 & $0.879 \pm 0.020$ & $0.967 \pm 0.003$ & $0.596 \pm 0.008$ \\
50 & 0.987 & $0.969 \pm 0.011$ & $0.986 \pm 0.002$ & $0.539 \pm 0.006$ \\
100 & 0.993 & $0.985 \pm 0.006$ & $0.995 \pm 0.001$ & $0.524 \pm 0.004$ \\
500 & 0.999 & $0.989 \pm 0.003$ & $0.999 \pm 0.001$ & $0.515 \pm 0.006$ \\
\bottomrule

  \end{tabular}
\end{table}

\begin{table}[ht]
  \centering
  \caption{WW domain (PF00397) $\rho$ sweep. $K=420$, $K_{\mathrm{des}}=69$.}
  \label{tab:ww-rho}
  \small
  \begin{tabular}{ccccc}
    \toprule
    $\rho$ & $f_{\mathrm{eff}}$ & $f_{\mathrm{obs}}$ & $\bar{a}_{\mathrm{des}}$ & $D$ \\
    \midrule
    1 & 0.164 & $0.144 \pm 0.016$ & $0.170 \pm 0.006$ & $0.686 \pm 0.004$ \\
2 & 0.282 & $0.161 \pm 0.009$ & $0.288 \pm 0.009$ & $0.688 \pm 0.005$ \\
5 & 0.496 & $0.187 \pm 0.021$ & $0.484 \pm 0.015$ & $0.686 \pm 0.003$ \\
10 & 0.663 & $0.223 \pm 0.028$ & $0.659 \pm 0.013$ & $0.670 \pm 0.007$ \\
20 & 0.797 & $0.255 \pm 0.027$ & $0.793 \pm 0.007$ & $0.671 \pm 0.003$ \\
50 & 0.908 & $0.299 \pm 0.025$ & $0.911 \pm 0.007$ & $0.655 \pm 0.003$ \\
100 & 0.952 & $0.291 \pm 0.019$ & $0.952 \pm 0.003$ & $0.652 \pm 0.004$ \\
500 & 0.990 & $0.330 \pm 0.016$ & $0.990 \pm 0.001$ & $0.631 \pm 0.004$ \\
\bottomrule

  \end{tabular}
\end{table}

\begin{table}[ht]
  \centering
  \caption{Homeobox domain (PF00046) $\rho$ sweep. $K=136$, $K_{\mathrm{des}}=102$.}
  \label{tab:homeobox-rho}
  \small
  \begin{tabular}{ccccc}
    \toprule
    $\rho$ & $f_{\mathrm{eff}}$ & $f_{\mathrm{obs}}$ & $\bar{a}_{\mathrm{des}}$ & $D$ \\
    \midrule
    1 & 0.750 & $0.931 \pm 0.017$ & $0.747 \pm 0.016$ & $0.523 \pm 0.005$ \\
2 & 0.857 & $0.939 \pm 0.007$ & $0.856 \pm 0.013$ & $0.518 \pm 0.002$ \\
5 & 0.938 & $0.953 \pm 0.008$ & $0.936 \pm 0.005$ & $0.519 \pm 0.003$ \\
10 & 0.968 & $0.948 \pm 0.004$ & $0.967 \pm 0.001$ & $0.520 \pm 0.003$ \\
20 & 0.984 & $0.953 \pm 0.014$ & $0.984 \pm 0.001$ & $0.519 \pm 0.005$ \\
50 & 0.993 & $0.954 \pm 0.015$ & $0.993 \pm 0.001$ & $0.519 \pm 0.004$ \\
100 & 0.997 & $0.927 \pm 0.015$ & $0.996 \pm 0.000$ & $0.535 \pm 0.006$ \\
500 & 0.999 & $0.945 \pm 0.011$ & $0.999 \pm 0.000$ & $0.536 \pm 0.005$ \\
\bottomrule

  \end{tabular}
\end{table}

\begin{table}[ht]
  \centering
  \caption{Forkhead domain (PF00250) $\rho$ sweep. $K=246$, $K_{\mathrm{des}}=122$.}
  \label{tab:forkhead-rho}
  \small
  \begin{tabular}{ccccc}
    \toprule
    $\rho$ & $f_{\mathrm{eff}}$ & $f_{\mathrm{obs}}$ & $\bar{a}_{\mathrm{des}}$ & $D$ \\
    \midrule
    1 & 0.496 & $0.565 \pm 0.013$ & $0.496 \pm 0.006$ & $0.485 \pm 0.003$ \\
2 & 0.663 & $0.590 \pm 0.031$ & $0.668 \pm 0.009$ & $0.485 \pm 0.007$ \\
5 & 0.831 & $0.625 \pm 0.025$ & $0.830 \pm 0.007$ & $0.469 \pm 0.004$ \\
10 & 0.908 & $0.612 \pm 0.019$ & $0.907 \pm 0.005$ & $0.468 \pm 0.003$ \\
20 & 0.952 & $0.609 \pm 0.030$ & $0.951 \pm 0.002$ & $0.484 \pm 0.003$ \\
50 & 0.980 & $0.688 \pm 0.013$ & $0.980 \pm 0.002$ & $0.434 \pm 0.004$ \\
100 & 0.990 & $0.696 \pm 0.031$ & $0.991 \pm 0.001$ & $0.415 \pm 0.002$ \\
500 & 0.998 & $0.731 \pm 0.028$ & $0.998 \pm 0.001$ & $0.402 \pm 0.004$ \\
\bottomrule

  \end{tabular}
\end{table}

\begin{table}[ht]
  \centering
  \caption{$\omega$-conotoxin O-superfamily $\rho$ sweep. $K=74$,
  $K_{\mathrm{des}}=23$.}
  \label{tab:conotoxin-rho}
  \small
  \begin{tabular}{ccccc}
    \toprule
    $\rho$ & $f_{\mathrm{eff}}$ & $f_{\mathrm{obs}}$ & $\bar{a}_{\mathrm{des}}$ & $D$ \\
    \midrule
    1 & 0.311 & $0.463 \pm 0.040$ & $0.304 \pm 0.025$ & $0.560 \pm 0.003$ \\
2 & 0.474 & $0.530 \pm 0.023$ & $0.467 \pm 0.010$ & $0.550 \pm 0.003$ \\
5 & 0.693 & $0.612 \pm 0.022$ & $0.667 \pm 0.021$ & $0.529 \pm 0.009$ \\
10 & 0.819 & $0.662 \pm 0.029$ & $0.819 \pm 0.013$ & $0.510 \pm 0.010$ \\
20 & 0.900 & $0.737 \pm 0.025$ & $0.899 \pm 0.019$ & $0.484 \pm 0.007$ \\
50 & 0.958 & $0.798 \pm 0.021$ & $0.958 \pm 0.004$ & $0.464 \pm 0.011$ \\
100 & 0.978 & $0.800 \pm 0.020$ & $0.982 \pm 0.002$ & $0.460 \pm 0.013$ \\
500 & 0.996 & $0.838 \pm 0.035$ & $0.996 \pm 0.002$ & $0.443 \pm 0.010$ \\
\bottomrule

  \end{tabular}
\end{table}

\begin{table}[ht]
  \centering
  \caption{$\beta$ sweep on Kunitz domain ($K=99$, $K_{\mathrm{des}}=32$).}
  \label{tab:beta-sweep}
  \small
  \begin{tabular}{cccc}
    \toprule
    $\rho$ & $\beta/\beta^{*}$ & $f_{\mathrm{obs}}$ & $D$ \\
    \midrule
    10  & 0.50 & 0.463 & 0.571 \\
10  & 1.00 & 0.487 & 0.537 \\
10  & 1.50 & 0.515 & 0.510 \\
10  & 2.00 & 0.534 & 0.496 \\
10  & 3.00 & 0.558 & 0.482 \\
\midrule
50  & 0.50 & 0.485 & 0.566 \\
50  & 1.00 & 0.535 & 0.529 \\
50  & 1.50 & 0.574 & 0.506 \\
50  & 2.00 & 0.606 & 0.495 \\
50  & 3.00 & 0.650 & 0.467 \\
\midrule
200 & 0.50 & 0.498 & 0.557 \\
200 & 1.00 & 0.556 & 0.518 \\
200 & 1.50 & 0.597 & 0.496 \\
200 & 2.00 & 0.631 & 0.478 \\
200 & 3.00 & 0.695 & 0.461 \\
\bottomrule

  \end{tabular}
\end{table}

\section{Matched Profile-HMM Sweep}
The matched sequence-weighted profile-HMM comparison was evaluated over the same multiplicity
grid as SA. Table~\ref{tab:profile-hmm-rho-si} reports four representative points; the
repository contains all eight multiplicity ratios, five independent replicates per condition,
and the designated-subset HMM endpoint. In the profile HMM, relative row weights act directly
on position-specific emission counts. Consequently, marker recovery approaches the weighted
input marginal more closely than it does after SA's PCA reconstruction and argmax decoder.

\begin{table}[p]
  \centering
  \caption{Matched SA and sequence-weighted profile-HMM benchmark across representative
  multiplicity ratios. Both methods use the same cleaned alignments, designation labels, and
  relative multiplicity ratio. Marker values are mean $\pm$ s.d. across five replicates of
  620 sequences; $D$ is mean pairwise sequence diversity.}
  \label{tab:profile-hmm-rho-si}
  \scriptsize
  \resizebox{\textwidth}{!}{%
  \begin{tabular}{lcccccc}
    \toprule
    Family & $\rho$ & $f_{\mathrm{eff}}$ & SA marker & HMM marker & SA $D$ & HMM $D$ \\
    \midrule
    Kunitz & 1 & 0.323 & $0.382 \pm 0.031$ & $0.306 \pm 0.018$ & 0.557 & 0.634 \\
Kunitz & 10 & 0.827 & $0.515 \pm 0.027$ & $0.820 \pm 0.016$ & 0.538 & 0.606 \\
Kunitz & 100 & 0.979 & $0.563 \pm 0.023$ & $0.980 \pm 0.006$ & 0.519 & 0.589 \\
Kunitz & 500 & 0.996 & $0.587 \pm 0.029$ & $0.996 \pm 0.002$ & 0.510 & 0.590 \\
\midrule
SH3 & 1 & 0.600 & $0.839 \pm 0.021$ & $0.545 \pm 0.013$ & 0.586 & 0.711 \\
SH3 & 10 & 0.938 & $0.900 \pm 0.012$ & $0.928 \pm 0.007$ & 0.581 & 0.665 \\
SH3 & 100 & 0.993 & $0.985 \pm 0.006$ & $0.989 \pm 0.002$ & 0.524 & 0.654 \\
SH3 & 500 & 0.999 & $0.989 \pm 0.003$ & $0.999 \pm 0.001$ & 0.515 & 0.655 \\
\midrule
WW & 1 & 0.164 & $0.144 \pm 0.016$ & $0.153 \pm 0.012$ & 0.686 & 0.677 \\
WW & 10 & 0.663 & $0.223 \pm 0.028$ & $0.662 \pm 0.013$ & 0.670 & 0.655 \\
WW & 100 & 0.952 & $0.291 \pm 0.019$ & $0.954 \pm 0.005$ & 0.652 & 0.630 \\
WW & 500 & 0.990 & $0.330 \pm 0.016$ & $0.988 \pm 0.004$ & 0.631 & 0.628 \\
\midrule
Homeobox & 1 & 0.750 & $0.931 \pm 0.017$ & $0.721 \pm 0.019$ & 0.523 & 0.643 \\
Homeobox & 10 & 0.968 & $0.948 \pm 0.004$ & $0.969 \pm 0.010$ & 0.520 & 0.613 \\
Homeobox & 100 & 0.997 & $0.927 \pm 0.015$ & $0.995 \pm 0.003$ & 0.535 & 0.605 \\
Homeobox & 500 & 0.999 & $0.945 \pm 0.011$ & $1.000 \pm 0.001$ & 0.536 & 0.605 \\
\midrule
Forkhead & 1 & 0.496 & $0.565 \pm 0.013$ & $0.492 \pm 0.022$ & 0.485 & 0.572 \\
Forkhead & 10 & 0.908 & $0.612 \pm 0.019$ & $0.902 \pm 0.011$ & 0.468 & 0.566 \\
Forkhead & 100 & 0.990 & $0.696 \pm 0.031$ & $0.991 \pm 0.004$ & 0.415 & 0.561 \\
Forkhead & 500 & 0.998 & $0.731 \pm 0.028$ & $0.997 \pm 0.001$ & 0.402 & 0.562 \\
\midrule
$\omega$-Conotoxin & 1 & 0.311 & $0.463 \pm 0.040$ & $0.320 \pm 0.019$ & 0.560 & 0.599 \\
$\omega$-Conotoxin & 10 & 0.819 & $0.662 \pm 0.029$ & $0.676 \pm 0.020$ & 0.510 & 0.469 \\
$\omega$-Conotoxin & 100 & 0.978 & $0.800 \pm 0.020$ & $0.802 \pm 0.014$ & 0.460 & 0.386 \\
$\omega$-Conotoxin & 500 & 0.996 & $0.838 \pm 0.035$ & $0.834 \pm 0.013$ & 0.443 & 0.381 \\
\bottomrule

  \end{tabular}
  }
\end{table}
\FloatBarrier

\section{Masking Compared with Multiplicity Weighting at the Same Temperature}
\label{app:matched-beta}
The hard mask and multiplicity weighting differ in one respect: the mask places the entire
equilibrium mixture on designated patterns, whereas a finite $\rho$ places a chosen share of it
there. Comparing the two at the same $\beta$ isolates that difference. We set $\rho$ so the
requested share was 0.5, 0.7, 0.9, 0.95, and 0.99 in turn, and ran the mask at each of the five
resulting temperatures on the Kunitz domain.

The decoded marker fractions barely separated. At a requested share of 0.5, the mask reached
0.510 and multiplicity weighting 0.425, a difference of 0.085 (SE 0.028), even though the two
conditions differed by 50 percentage points in what they requested. That difference shrank to
0.053 (SE 0.029) at a requested share of 0.7, 0.025 at 0.9, 0.012 at 0.95, and 0.003
(SE 0.030) at 0.99, where the two requests nearly coincide. Doubling the requested designated
mass therefore bought less than a tenth of a unit of decoded marker. This result
again shows compression across reconstruction and discrete decoding.

\section{Cav2.2 Complex-Prediction Diagnostic}
We modeled SA-generated and natural $\omega$-conotoxins with the Cav2.2 pore-domain target
using ColabFold AlphaFold2-multimer. This calculation was treated as a diagnostic rather than
as validation of the generation method. The SA designated-subset group ($n=10$) gave pTM
$0.221\pm0.007$, iPTM $0.105\pm0.008$, and interface pLDDT $37.2\pm5.4$; the SA full-family
group ($n=10$) gave $0.219\pm0.008$, $0.104\pm0.007$, and $38.2\pm4.0$; and natural controls
($n=5$) gave $0.228\pm0.004$, $0.108\pm0.007$, and $40.7\pm3.5$. The uniformly low scores,
including those for natural controls, show that this setup did not resolve a reliable complex
geometry.

\begin{table}[ht]
  \centering
  \footnotesize
  \setlength{\tabcolsep}{3.5pt}
  \caption{Low-confidence AlphaFold2-multimer predictions for generated and control
  $\omega$-conotoxins. Values are group means $\pm$ s.d.}
  \label{tab:docking-validation}
  \begin{tabular}{lcccc}
    \toprule
    Group & $n$ & Mean pTM & Mean iPTM & Mean interface pLDDT \\
    \midrule
    SA designated-subset & 10 & $0.221\pm0.007$ & $0.105\pm0.008$ & $37.2\pm5.4$ \\
    SA full-family & 10 & $0.219\pm0.008$ & $0.104\pm0.007$ & $38.2\pm4.0$ \\
    Natural controls & 5 & $0.228\pm0.004$ & $0.108\pm0.007$ & $40.7\pm3.5$ \\
    \bottomrule
  \end{tabular}
\end{table}

Group-level comparisons among the SA designated-subset, SA full-family, and
natural $\omega$-conotoxin controls were evaluated using exact two-sided
permutation tests on mean complex-level scores (iPTM, pTM, confidence, and
interface pLDDT), with sample sizes 10, 10, and 5, respectively. No pairwise
contrast reached significance at $\alpha=0.05$ (all $p > 0.05$; smallest observed
$p=0.0726$ for confidence, SA full vs controls).

\begin{table}[ht]
  \centering
  \caption{Pairwise exact permutation p-values for $\omega$-conotoxin docking comparisons (all $H_0$: equal means).}
  \label{tab:docking-permutation-pvalues}
  \resizebox{\textwidth}{!}{%
  \begin{tabular}{lccc}
    \toprule
    \textbf{Metric} & \textbf{SA des. vs SA full} & \textbf{SA des. vs controls} & \textbf{SA full vs controls} \\
    \midrule
    iPTM & 1.0000 & 0.7502 & 0.4752 \\
    pTM & 0.7873 & 0.1285 & 0.1092 \\
    Confidence & 0.6560 & 0.1245 & 0.0726 \\
    Interface pLDDT & 0.6556 & 0.2238 & 0.2934 \\
    \bottomrule
  \end{tabular}
  }
\end{table}

\section{The Boltzmann target is an exact Gaussian mixture}\label{app:gmm}

The sampler state $\boldsymbol{\xi}$ is a free point in $\mathbb{R}^{d}$, where $d$
is the number of PCA coordinates. Each of the $K$ stored memories
$\mathbf{m}_1,\ldots,\mathbf{m}_K$ has length one and carries a multiplicity weight
$r_k > 0$. The state itself has no fixed length, because nothing in the update
rescales it back onto the unit sphere. It can land anywhere, which is why the answer
below is a distribution over all of $\mathbb{R}^{d}$.
Where does the sampler settle if we run it for a long time? For the underlying
continuous-time dynamics the answer is the Boltzmann distribution
$p_\beta(\boldsymbol{\xi})\propto e^{-\beta E_r(\boldsymbol{\xi})}$, built from the
multiplicity-weighted Hopfield energy:
\begin{equation}
  E_r(\boldsymbol{\xi})
  = \tfrac12\lVert\boldsymbol{\xi}\rVert^2
    - \tfrac1\beta\log\sum_{k=1}^{K} r_k\, e^{\beta\, \mathbf{m}_k^\top\boldsymbol{\xi}},
  \qquad \lVert\mathbf{m}_k\rVert = 1 .
\end{equation}
This section shows that distribution is exactly a mixture of Gaussians. The update
we actually run, Eq.~\MainEqRef{eq:weighted-ula}, takes finite steps, so it only
comes close to this distribution rather than landing on it
exactly~\cite{durmusMoulinesULA2017}.

The proof takes three steps. First we exponentiate, which clears the logarithm and
leaves one sum over the memories. Then we complete the square, which puts each term
in the shape of a Gaussian. Then we normalize, which fixes the constants and gives
the mixture weights.

It helps to write down the shape we are aiming for first. A Gaussian in $d$
dimensions with mean $\boldsymbol{\mu}$ and the same variance $\sigma^{2}$ in every
direction has density:
\begin{align}
  \mathcal{N}(\boldsymbol{\xi};\,\boldsymbol{\mu},\,\sigma^{2}\mathbf{I})
  &= \left(2\pi\sigma^{2}\right)^{-d/2}
     \exp\!\left[-\frac{1}{2\sigma^{2}}
       \lVert\boldsymbol{\xi}-\boldsymbol{\mu}\rVert^{2}\right]
     \notag\\[4pt]
  \mathcal{N}(\boldsymbol{\xi};\,\boldsymbol{\mu},\,\beta^{-1}\mathbf{I})
  &= \left(\tfrac{\beta}{2\pi}\right)^{\!d/2}
     \exp\!\left[-\tfrac\beta2
       \lVert\boldsymbol{\xi}-\boldsymbol{\mu}\rVert^{2}\right] .
  \label{eq:si-normal}
\end{align}
The second line is the first with $\sigma^{2}$ set to $\beta^{-1}$, the value we
need below. The only place $\boldsymbol{\xi}$ appears is inside
$\lVert\boldsymbol{\xi}-\boldsymbol{\mu}\rVert^{2}$, its squared distance to the
mean. So whenever we see something of the form
$\exp[-\tfrac\beta2\lVert\boldsymbol{\xi}-\mathbf{c}\rVert^{2}]$, we are looking at
a Gaussian centered at $\mathbf{c}$ with variance $\beta^{-1}$, apart from the
constant $(\beta/2\pi)^{d/2}$ in front of it. Everything below is getting each term
into that form with $\mathbf{c} = \mathbf{m}_k$. That is what makes each memory the
center of a Gaussian, and $\beta$ the thing that sets its width.

\paragraph{Exponentiation.}
Multiply the energy by $-\beta$. That $-\beta$ cancels the $-\tfrac1\beta$ sitting
in front of the logarithm, so the log term is left with nothing multiplying it:
\begin{align}
  -\beta E_r(\boldsymbol{\xi})
  &= -\beta\left[\tfrac12\lVert\boldsymbol{\xi}\rVert^2
     - \tfrac1\beta\log\sum_{k=1}^{K} r_k\,
       e^{\beta\, \mathbf{m}_k^\top\boldsymbol{\xi}}\right]
     \notag\\[2pt]
  &= -\tfrac\beta2\lVert\boldsymbol{\xi}\rVert^2
     + \log\sum_{k=1}^{K} r_k\, e^{\beta\, \mathbf{m}_k^\top\boldsymbol{\xi}} .
\end{align}
Now raise $e$ to both sides. A sum in the exponent becomes a product of two
factors, and $e^{\log S} = S$ cancels the logarithm in the second one, which frees
the sum:
\begin{align}
  e^{-\beta E_r(\boldsymbol{\xi})}
  &= e^{-\frac\beta2\lVert\boldsymbol{\xi}\rVert^2}\;
     \exp\!\left[\log\sum_{k=1}^{K} r_k\,
       e^{\beta\, \mathbf{m}_k^\top\boldsymbol{\xi}}\right]
     \notag\\[2pt]
  &= e^{-\frac\beta2\lVert\boldsymbol{\xi}\rVert^2}
     \sum_{k=1}^{K} r_k\, e^{\beta\, \mathbf{m}_k^\top\boldsymbol{\xi}}
     \notag\\[2pt]
  &= \sum_{k=1}^{K} r_k\,
     \exp\!\left[-\tfrac\beta2\lVert\boldsymbol{\xi}\rVert^2
       + \beta\, \mathbf{m}_k^\top\boldsymbol{\xi}\right] .
\end{align}
The last line pulls the $e^{-\frac\beta2\lVert\boldsymbol{\xi}\rVert^{2}}$ factor
inside the sum, which is fine because it does not depend on $k$. From here we work
on one term at a time, and the job is to make its exponent look like the one in
Eq.~\ref{eq:si-normal}.

\paragraph{Completing the square.}
We want the exponent
$-\tfrac\beta2\lVert\boldsymbol{\xi}\rVert^{2}
 + \beta\,\mathbf{m}_k^\top\boldsymbol{\xi}$
to look like $-\tfrac\beta2\lVert\boldsymbol{\xi}-\mathbf{m}_k\rVert^{2}$. It almost
does. Both start with $-\tfrac\beta2\lVert\boldsymbol{\xi}\rVert^{2}$, but in ours
the memory sits outside a square instead of inside one. The quickest way to see the
difference is to start from what we want and expand it:
\begin{align}
  -\tfrac\beta2\lVert\boldsymbol{\xi}-\mathbf{m}_k\rVert^2
  &= -\tfrac\beta2\left[\lVert\boldsymbol{\xi}\rVert^2
     - 2\,\mathbf{m}_k^\top\boldsymbol{\xi}
     + \lVert\mathbf{m}_k\rVert^2\right]
     \notag\\[2pt]
  &= \underbrace{-\tfrac\beta2\lVert\boldsymbol{\xi}\rVert^2
     + \beta\,\mathbf{m}_k^\top\boldsymbol{\xi}}_{\text{exponent of the }k\text{th term}}
     \;-\; \tfrac\beta2\lVert\mathbf{m}_k\rVert^2 .
\end{align}
The underbraced part is our exponent. So our exponent equals
$-\tfrac\beta2\lVert\boldsymbol{\xi}-\mathbf{m}_k\rVert^{2}
 + \tfrac\beta2\lVert\mathbf{m}_k\rVert^{2}$, which is the shape we want plus a
constant. Putting that into every term, then splitting the constant off as its own
factor:
\begin{align}
  e^{-\beta E_r(\boldsymbol{\xi})}
  &= \sum_{k=1}^{K} r_k\,
     \exp\!\left[-\tfrac\beta2\lVert\boldsymbol{\xi}-\mathbf{m}_k\rVert^2
       + \tfrac\beta2\lVert\mathbf{m}_k\rVert^2\right]
     \notag\\[2pt]
  &= \sum_{k=1}^{K} r_k\, e^{\frac\beta2\lVert\mathbf{m}_k\rVert^2}\,
     e^{-\frac\beta2\lVert\boldsymbol{\xi}-\mathbf{m}_k\rVert^2} .
\end{align}
The last factor is now the Gaussian shape from Eq.~\ref{eq:si-normal}, with
$\mathbf{c} = \mathbf{m}_k$. So term $k$ is a Gaussian centered on memory $k$ with
variance $\beta^{-1}$: a large $\beta$ makes it a tight bump around that memory, a
small $\beta$ a broad one. All that is left is the constants.

\paragraph{Unit norm and normalization.}
This is the step where unit norm matters. Because $\lVert\mathbf{m}_k\rVert = 1$
for every $k$, the leftover constant $e^{\frac\beta2\lVert\mathbf{m}_k\rVert^2}$ is
the same number $e^{\beta/2}$ in every term, so it comes out of the sum. Each
remaining
factor still needs the constant $(\beta/2\pi)^{d/2}$ to be a full Gaussian density,
so we put it in and divide it back out, which is the $(2\pi/\beta)^{d/2}$ in front:
\begin{align}
  e^{-\beta E_r(\boldsymbol{\xi})}
  &= e^{\beta/2}\sum_{k=1}^{K} r_k\,
     e^{-\frac\beta2\lVert\boldsymbol{\xi}-\mathbf{m}_k\rVert^2}
     \notag\\[2pt]
  &= e^{\beta/2}\left(\tfrac{2\pi}{\beta}\right)^{\!d/2}
     \sum_{k=1}^{K} r_k\,
     \mathcal{N}(\boldsymbol{\xi};\,\mathbf{m}_k,\,\beta^{-1}\mathbf{I}) .
\end{align}
To turn the $\propto$ into an $=$, we need the total $Z$, the integral over all of
$\mathbb{R}^d$. Every Gaussian density integrates to one, so the integral leaves
nothing behind but the sum of the $r_j$:
\begin{align}
  Z &= \int_{\mathbb{R}^d} e^{-\beta E_r(\boldsymbol{\xi})}\,
       \mathrm{d}\boldsymbol{\xi}
     \notag\\[2pt]
    &= e^{\beta/2}\left(\tfrac{2\pi}{\beta}\right)^{\!d/2}
       \sum_{j=1}^{K} r_j
       \underbrace{\int_{\mathbb{R}^d}
         \mathcal{N}(\boldsymbol{\xi};\,\mathbf{m}_j,\,\beta^{-1}\mathbf{I})\,
         \mathrm{d}\boldsymbol{\xi}}_{=\;1}
     \notag\\[2pt]
    &= e^{\beta/2}\left(\tfrac{2\pi}{\beta}\right)^{\!d/2}
       \sum_{j=1}^{K} r_j .
\end{align}
Dividing one by the other, $e^{\beta/2}$ and $(2\pi/\beta)^{d/2}$ sit on both the
top and the bottom, so they cancel and never matter:
\begin{align}
  p_\beta(\boldsymbol{\xi})
  &= \frac{e^{-\beta E_r(\boldsymbol{\xi})}}{Z}
   = \frac{e^{\beta/2}\left(\tfrac{2\pi}{\beta}\right)^{\!d/2}
           \sum_{k=1}^{K} r_k\,
           \mathcal{N}(\boldsymbol{\xi};\,\mathbf{m}_k,\,\beta^{-1}\mathbf{I})}
          {e^{\beta/2}\left(\tfrac{2\pi}{\beta}\right)^{\!d/2}
           \sum_{j=1}^{K} r_j}
     \notag\\[4pt]
  &= \sum_{k=1}^{K}
     \frac{r_k}{\sum_{j=1}^{K} r_j}\,
     \mathcal{N}(\boldsymbol{\xi};\,\mathbf{m}_k,\,\beta^{-1}\mathbf{I}) .
\end{align}
Writing $w_k = r_k / \sum_j r_j$ for the normalized multiplicities, this is given by:
\begin{equation}
  \boxed{\;
  p_\beta(\boldsymbol{\xi})
  = \sum_{k=1}^{K} w_k\, \mathcal{N}(\boldsymbol{\xi};\,\mathbf{m}_k,\,\beta^{-1}\mathbf{I}),
  \qquad w_k = \frac{r_k}{\sum_j r_j} \; }
\end{equation}
an exact mixture of Gaussians, one per stored memory, each centered on its own
memory with variance $\beta^{-1}$, and weighted by its multiplicity divided by the
total. If the memories had not been unit norm, the
$e^{\frac\beta2\lVert\mathbf{m}_k\rVert^2}$ would differ from term to term and would
not come out of the sum, so the weights would be
$r_k\,e^{\frac\beta2\lVert\mathbf{m}_k\rVert^2}$ instead. The result would still be
a Gaussian mixture, just not with these weights.

\paragraph{Corollaries.}
\emph{(i) Exact sampling.} Draw $k\sim\mathrm{Categorical}(w)$ and
$\boldsymbol{\xi}=\mathbf{m}_k+\beta^{-1/2}\mathbf{z}$, $\mathbf{z}\sim\mathcal{N}(\mathbf{0},\mathbf{I})$;
no Markov chain is required. The continuous-time Langevin diffusion has
$p_\beta$ as its stationary law, while finite-step ULA is subject to
discretization and finite-time error.
\emph{(ii) Exact designated control.} Let the designated subset $B$ hold
$K_{\mathrm{des}}$ of the $K$ patterns, with the remaining
$K_{\mathrm{bg}} = K - K_{\mathrm{des}}$ forming the background. The probability
mass on $B$ is $\sum_{k\in B} w_k = \sum_{k\in B} r_k / \sum_j r_j
\equiv f_{\mathrm{eff}}$. Setting $r_k=\rho$ on $B$ and $1$ otherwise gives
$f_{\mathrm{eff}} = \rho K_{\mathrm{des}} / (\rho K_{\mathrm{des}} +
K_{\mathrm{bg}})$, with no equal-similarity assumption.
\emph{(iii) Score.} $\nabla_{\boldsymbol{\xi}}\log p_\beta
= -\beta\nabla E_r
= \beta\bigl[\mathbf{X}\,\mathrm{softmax}(\beta\mathbf{X}^\top\boldsymbol{\xi}+\log\mathbf{r})
  - \boldsymbol{\xi}\bigr]$, whose Euler-Maruyama discretization is
Eq.~\MainEqRef{eq:weighted-ula}.

\paragraph{Exact-equilibrium benchmark.}
We compared the finite-step ULA implementation with independent draws from
Eq.~\MainEqRef{eq:gmm-target} on the Kunitz memory. For each multiplicity ratio,
1{,}640 exact and 1{,}640 ULA states were decoded through the same PCA inverse
map and residue-wise argmax. The analytic designated-component probability
$f_{\mathrm{eff}}$ is reported separately from MAP-designated basin occupancy,
a hard assignment based on the maximum-posterior component
(Table~\ref{tab:gmm-benchmark}). The latter need not equal the mixture weight when
Gaussian components overlap.

\begin{table}[ht]
  \centering
  \caption{Exact Gaussian-mixture versus finite-step ULA sampling on Kunitz.
  Exact and ULA columns use equal sample counts and the same $\beta^{*}$.
  The ``MAP des.'' column is the fraction assigned to a designated
  maximum-posterior component and is not the latent mixture weight
  $f_{\mathrm{eff}}$. The AA KL column is the amino acid composition
  divergence between the decoded exact and ULA libraries.}
  \label{tab:gmm-benchmark}
  \small
  \resizebox{\textwidth}{!}{%
  \begin{tabular}{cccccccc}
    \toprule
    $\rho$ & $\beta^{*}$ & $f_{\mathrm{eff}}$
      & MAP des., exact & MAP des., ULA
      & P1 K/R, exact & P1 K/R, ULA & AA KL \\
    \midrule
    1   & 4.27 & 0.323 & 0.307 & 0.281 & 0.398 & 0.385 & $1.40\times10^{-4}$ \\
    10  & 4.93 & 0.827 & 0.894 & 0.887 & 0.515 & 0.499 & $1.25\times10^{-4}$ \\
    500 & 8.78 & 0.996 & 0.997 & 0.999 & 0.604 & 0.597 & $1.19\times10^{-4}$ \\
    \bottomrule
  \end{tabular}
  }
\end{table}

\end{document}